\documentclass[usenames,dvipsnames]{article}
\usepackage{microtype}
\usepackage{graphicx}
\usepackage{subcaption}
\usepackage{booktabs} 
\usepackage{tikz} 
\usepackage{tikz-layers} 
\usepackage{xcolor} 
\usepackage{amsmath} 
\usepackage{amssymb} 
\usepackage{wasysym} 
\usepackage{array}
\newcommand{\PreserveBackslash}[1]{\let\temp=\\#1\let\\=\temp}
\newcolumntype{C}[1]{>{\PreserveBackslash\centering}p{#1}}
\newcolumntype{R}[1]{>{\PreserveBackslash\raggedleft}p{#1}}
\newcolumntype{L}[1]{>{\PreserveBackslash\raggedright}p{#1}}

\usepackage{hyperref}

\usepackage[accepted]{icml2022}

\providecommand\sD{\ensuremath{\mathcal{D}}}

\providecommand\sL{\ensuremath{\mathcal{L}}}

\newcommand\eqdef{\ensuremath{\stackrel{\rm def}{=}}} 

\newcommand\refsec[1]{Section~\ref{sec:#1}}

\newcommand\reffig[1]{Figure~\ref{fig:#1}}

\newcommand\reftab[1]{Table~\ref{tab:#1}}
\newcommand\refapp[1]{Appendix~\ref{sec:#1}}

\ifthenelse{\isundefined{\definition}}{\newtheorem{definition}{Definition}}{}
\ifthenelse{\isundefined{\assumption}}{\newtheorem{assumption}{Assumption}}{}
\ifthenelse{\isundefined{\hypothesis}}{\newtheorem{hypothesis}{Hypothesis}}{}
\ifthenelse{\isundefined{\proposition}}{\newtheorem{proposition}{Proposition}}{}
\ifthenelse{\isundefined{\theorem}}{\newtheorem{theorem}{Theorem}}{}
\ifthenelse{\isundefined{\lemma}}{\newtheorem{lemma}{Lemma}}{}
\ifthenelse{\isundefined{\corollary}}{\newtheorem{corollary}{Corollary}}{}
\ifthenelse{\isundefined{\alg}}{\newtheorem{alg}{Algorithm}}{}
\ifthenelse{\isundefined{\example}}{\newtheorem{example}{Example}}{}
\newcommand{\E}{\ensuremath{\mathbb{E}}} 
\newcommand{\inpainter}{dialog inpainter}
\newcommand{\di}[0]{dialog inpainting}

\newcommand{\Concat}{\operatorname{Concat}}

\newcommand{\embed}{\operatorname{embed}}

\newcommand{\Inpaint}[0]{\ensuremath{\operatorname{\text{Inpaint}}}}
\newcommand{\InpaintST}[0]{\ensuremath{\Inpaint_{\text{PT}}}}
\newcommand{\InpaintOQ}[0]{\ensuremath{\Inpaint_{\text{OQ}}}}
\newcommand{\InpaintSTOQ}[0]{\ensuremath{\Inpaint_{\text{PTOQ}}}}

\newcommand{\WikiDialog}[0]{\textrm{WikiDialog}}
\newcommand{\WikiDialogST}[0]{\ensuremath{\WikiDialog_{\text{PT}}}}
\newcommand{\WikiDialogOQ}[0]{\ensuremath{\WikiDialog_{\text{OQ}}}}
\newcommand{\WikiDialogSTOQ}[0]{\ensuremath{\WikiDialog_{\text{PTOQ}}}}
\newcommand{\WebDialog}[0]{\textrm{WebDialog}}
\newcommand{\WebDialogPT}[0]{\ensuremath{\WebDialog_{\text{PT}}}}

\newcommand{\TextWikiDialog}[0]{\text{WikiDialog}}
\newcommand{\TextWikiDialogST}[0]{\ensuremath{\TextWikiDialog_{\text{PT}}}}
\newcommand{\TextWikiDialogOQ}[0]{\ensuremath{\TextWikiDialog_{\text{OQ}}}}
\newcommand{\TextWikiDialogSTOQ}[0]{\ensuremath{\TextWikiDialog_{\text{PTOQ}}}}
\newcommand{\TextWebDialog}[0]{\text{WebDialog}}

\newcommand{\qrecc}[0]{\text{QReCC}}
\newcommand{\orquac}[0]{\text{OR-QuAC}}

\newcommand{\cast}[0]{\text{CAsT}}
\newcommand{\msmarco}[0]{\text{MS Marco}}
\newcommand{\paq}[0]{\text{PAQ}}

\newcommand{\socialmedia}[0]{\textrm{PublicDialog}}
\newcommand{\taskmaster}[0]{\textrm{TaskMaster}}

\newcommand{\qreccb}[0]{\textrm{\textbf{Q}ReCC}}
\newcommand{\orquacb}[0]{\textrm{\textbf{O}R-QuAC}}
\newcommand{\socialmediab}[0]{\textrm{\textbf{P}ublicDialog}}
\newcommand{\taskmasterb}[0]{\text{\textbf{T}askMaster}}

\usetikzlibrary{calc}
\usetikzlibrary{fit}
\usetikzlibrary{patterns}

\usepackage{pifont}
\icmltitlerunning{Dialog Inpainting: Turning Documents into Dialogs
}

\begin{document}

\twocolumn[
\icmltitle{Dialog Inpainting: Turning Documents into Dialogs}



\icmlsetsymbol{equal}{*}

\begin{icmlauthorlist}
\icmlauthor{Zhuyun Dai}{equal,google}
\icmlauthor{Arun Tejasvi Chaganty}{equal,google}
\icmlauthor{Vincent Zhao}{equal,google}
\icmlauthor{Aida Amini}{google}
\icmlauthor{Qazi Mamunur Rashid}{google}
\icmlauthor{Mike Green}{google}
\icmlauthor{Kelvin Guu}{equal,google}
\end{icmlauthorlist}

\icmlaffiliation{google}{Google Inc., Mountain View, USA}

\icmlcorrespondingauthor{Dialog Inpainting Core}{dialog-inpainting-core@google.com}

\icmlkeywords{Large language models, question answering, conversations, ICML}

\vskip 0.3in
]



\printAffiliationsAndNotice{\icmlEqualContribution} 

\begin{abstract}
Many important questions (e.g. ``How to eat healthier?'') require conversation to establish context and explore in depth.
However, conversational question answering (ConvQA) systems have long been stymied by scarce training data that is expensive to collect.
To address this problem, we propose a new technique for synthetically generating diverse and high-quality dialog data: \emph{dialog inpainting}.
Our approach takes the text of any document and transforms it into a two-person dialog between the writer and an imagined reader:
    we treat sentences from the article as utterances spoken by the writer, and then use a dialog inpainter to predict what the imagined reader asked or said in between each of the writer's utterances. 
By applying this approach to passages from Wikipedia and the web, we produce \WikiDialog{} and \WebDialog{}, two datasets totalling 19 million diverse information-seeking dialogs---1,000x larger than the largest existing ConvQA dataset.
Furthermore, human raters judge the \emph{answer adequacy} and \emph{conversationality} of \WikiDialog{} to be as good or better than existing manually-collected datasets.
Using our inpainted data to pre-train ConvQA retrieval systems, we significantly advance state-of-the-art across three benchmarks (\qrecc{}, \orquac{}, TREC \cast{}) yielding up to 40\% relative gains on standard evaluation metrics.
\end{abstract}
\section{Introduction}
\label{sec:intro}

\begin{figure}[t]
\centering
\includegraphics[width=0.8\columnwidth]{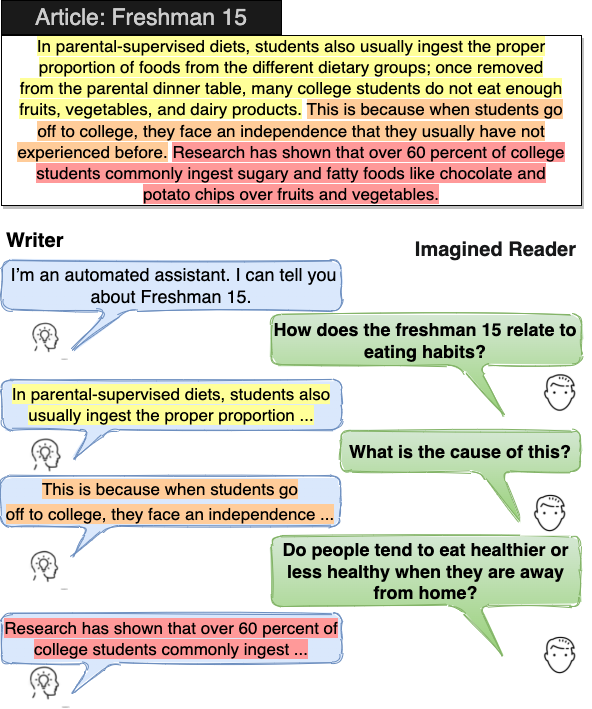}
\caption{\label{fig:document2dialog-example}
A real example of a dialog inferred from a Wikipedia passage using dialog inpainting. Highlighted utterances are original sentences from the article.
All other utterances are generated by the \inpainter{}.
}
\end{figure}

Modern information-seeking tools such as web search and question answering~\citep{Karpukhin2020DPR,zhu2021retrieving} excel at questions that have well-defined answers (e.g., ``Where was Barack Obama born?'').
But many important questions are more open-ended---e.g., ``How to eat healthier?''---and require conversation to elicit context and explore in depth:
    ``How do I eat more protein?'',
    ``What about vegetarians?''.
\emph{Conversational} question answering systems (ConvQA)~\citep{stede2004information,radlinski2017theoretical,Culpepper0S18}, would empower users to answer these questions as if they could discuss with an expert at any time.

Despite this promising vision, progress has been stymied by scarce training data.
While conversational data is abundant in online forums, much of it focuses on personal anecdotes and subjective opinions,
    and is thus unsuitable for an information-seeking system:
    we desire responses that minimize personal biases and cite reliable sources.
Directly crowdsourcing dialogs is also hard:
    crowdworkers are rarely experts in the domain of interest and tend to overlook important questions or provide shallow answers~\citep{li2021ditch}.
It is also expensive: the largest extant datasets contain only about 10,000 conversations each~\citep{Choi2018QuAC,reddy2018coqa,Dinan2018wow,Saeidi2018sharc,Campos2020doqa,Feng2020doc2dial,Anantha2021QReCC}.

On the other hand, high-quality documents, such as those in Wikipedia or PubMed, are abundant.
These documents are often edited or written by experts who have invested significant time streamlining their discourse and anticipating a reader's questions.
What if we could rewrite these documents into dialogs between the writer and their imagined reader?
This would yield an enormous corpus of information-seeking dialogs with attributable, expert answers that could then be used to train a ConvQA system.
We aim to achieve this with dialog inpainting.

To transform any document into a dialog, our first observation is that we already know what the writer wants to discuss---that is reflected in the original text of the document.
So, we pretend that the original document is the transcript of what the writer said to the reader in an imagined dialog.
But we are still missing what the reader asked.
This is like overhearing someone else's phone call: you hear one side, but not the other.
Oftentimes, one can still guess what the other side was saying --- we call this prediction task \emph{\di{}}, because we are ``inpainting'' the missing parts of the dialog that we did not hear (inspired by the term's usage in computer vision \cite{IizukaS017,LiuRSWTC18,Yu0YSLH18}).
Drawing on this intuition,
    we train an \emph{inpainter} model to predict missing utterances in a dialog,
    and use it to predict the unobserved questions in a document.
    By interleaving the generated questions and sentences from the document, we form a dialog (\reffig{document2dialog-example}).

We apply our inpainter to passages from Wikipedia and the web, yielding \WikiDialog{} and \WebDialog{},\footnote{%
     We released \WikiDialog{} at \url{https://github.com/google-research/dialog-inpainting}}
two datasets totalling 19M+ dialogs --- 1,000x larger than the largest existing ConvQA dataset.
When evaluated for \emph{conversationality} and \emph{answer adequacy}, we surprisingly find that our synthetically generated data is as good or better than previous crowd-sourced datasets (\refsec{dataset}).
Our generated dialogs inherit the good qualities of the professionally written documents we inpaint (topical diversity, coherent discourse, evidence-backed claims, etc.) without needing to train on dialog data of the same quality.



Importantly, we find that our inpainted datasets are powerful sources of training data for ConvQA systems (\refsec{application}).
When used to pre-train standard retriever and reranker architectures,
    they advance state-of-the-art across three different ConvQA retrieval benchmarks (QRECC, OR-QUAC, TREC-CAST),
    delivering up to 40\% relative gains on standard evaluation metrics (\refsec{evaluation}).
Remarkably, we find that just pre-training on WikiDialog enables strong zero-shot retrieval performance---up to 95\% of a finetuned retriever's performance---without using any in-domain ConvQA data.
\section{Dialog Inpainting}
\label{sec:method}


\label{sec:di_task}
The goal of dialog inpainting is to take a \emph{partial dialog} (one where some of the speaker turns are unobserved), and generate a \emph{complete dialog} (one where all unobserved turns have been filled in with the model's predictions).

Formally, a complete dialog $d$ is a sequence of speaker utterances, $d = (u_1, u_2, \ldots, u_t, \ldots, u_T)$.
We use the same notation for partial dialogs, denoting unobserved utterances with the $\diamond$ symbol.
For example, $(u_1, u_2, \diamond, u_4, \diamond)$ is a partial dialog where utterances $u_3$ and $u_5$ are unobserved. We refer to these as ``masked'' utterances.
We also use the shorthand $d_{m(3, 5)}$ to denote a dialog $d$ with utterances 3 and 5 masked.

To complete the partial dialog $d_{m(3, 5)}$, we generate predictions for utterances 3 and 5, denoted $\hat{u}_3$ and $\hat{u}_5$.
The inpainted dialog is then: 
\begin{equation*}
\Inpaint(d_{m(3, 5)}) = (u_1, u_2, \hat{u}_3, u_4, \hat{u_5}).
\end{equation*}

\subsection{Training: Dialog reconstruction}
\label{sec:di_train}
We train a \emph{\inpainter{}} using the following dialog reconstruction task:
Given a complete dialog, $d = (u_1, u_2, \ldots, u_T)$, we randomly mask one utterance, $u_t$, yielding a partial dialog:
\begin{equation*}
d_{m(t)} = (u_1, \ldots, u_{t-1}, \diamond, u_{t+1}, \ldots, u_T).
\end{equation*}
Given this partial dialog, we train our model to predict $u_t$, the original value of the masked utterance. This is similar to the masked language modeling task used by BERT~\cite{Devlin2019BERT}, but instead of reconstructing missing tokens in a passage, we reconstruct a missing utterance in a dialog.

More precisely, let us assume that the \inpainter{} is a generative model with parameters $\theta$ specifying a probability distribution $p_\theta(u_t \mid d_{m(t)})$. Then, our training objective is to minimize the following loss:
\begin{equation}
\sL(\theta) = -\sum_{d \in \sD} \E_{u_t \sim d}[\log p_\theta(u_t \mid d_{m(t)})],
\label{eq:train_objective}
\end{equation}
where $\sD$ is a corpus of complete dialogs and $u_t$ is a randomly sampled utterance from the dialog $d$.

We implement our inpainter using the T5 \citep{Raffel2020T5} text-to-text encoder-decoder Transformer~\cite{vaswani2017attention},
    where the input $x$ and output $y$ are represented as text strings.
We convert each dialog $d$ into an $(x, y)$ pair as follows:
First, we randomly sample a turn $t$ to mask, yielding partial dialog $d_{m(t)}$ and target output $u_t$.
To form input string $x$,
    we prepend each utterance in $d_{m(t)}$ with its corresponding speaker id (e.g., 0 or 1 in a two-speaker dialog),
    and simply concatenate them together, for example, ``\texttt{0:$u_1$ 1:$u_2$ 0:$\diamond$ 1:${u_4}$}''.
The output string $y$ is simply the masked utterance $u_t$.
We optimize this model using a standard cross-entropy loss which is equivalent to Eq.~\ref{eq:train_objective}.


\subsection{Inference: Transforming documents into dialogs}
\label{sec:di_inference_details}
We now show how to use a trained inpainter to transform a document into a dialog. Suppose we have a document or passage $p$ consisting of sentences $(s_1, s_2, \ldots, s_m)$.
Imagine that each of these sentences was an utterance spoken by the writer in an imagined dialog with the reader:
    we want to know what the imagined reader said between each of the writer's utterances.
We can pose this question to our inpainter by asking it to complete the following partial dialog:
    $(\diamond, s_1, \diamond, s_2, \diamond, \ldots, \diamond, s_m)$---each utterance from the imagined reader starts masked and is responded to by the writer with a sentence from the passage.
While the model knows the masked utterances come from another speaker, it still lacks any hint that the speaker should ask questions.
To provide this hint, we prepend an utterance from the writer $s_\text{prompt}$ that says: 
\textit{``Hello, I am an automated assistant and can answer questions about \texttt{(document title)}''}.
Hence, the overall partial dialog has the form:
\begin{equation*}
\operatorname{PartialDialog}(p) = (s_\text{prompt}, \diamond, s_1, \diamond, \ldots, \diamond, s_m).
\end{equation*}

Note, however, that this partial dialog contains multiple masked utterances, while our inpainter is only trained to inpaint a single utterance at a time.
To handle this, we use the model autoregressively:
we begin by providing $(s_\text{prompt}, \diamond, s_1)$ as input to the inpainter and generate $\hat{u}_1$ via greedy decoding.
We then replace the first mask with $\hat{u}_1$ and use a new input $(s_\text{prompt}, \hat{u}_1, s_1, \diamond, s_2)$ to generate $\hat{u}_2$
    and so on until all masks are filled and the dialog is complete.

\subsection{Case study: Applying dialog inpainting to generate an information seeking dialog dataset}
\label{sec:di_train_details}
We now have all the ingredients necessary to generate synthetic information-seeking dialogs at scale.
In this section, we will outline how we trained inpainters and applied them on two document corpora to respectively generate two such datasets: \WikiDialog{} and \WebDialog{}.

\textbf{Inpainter model training.}
We used \socialmediab{}, \taskmasterb{}, \orquacb{}, and \qreccb{} (\reftab{inpainter-data}) to train three different inpainters: \InpaintST{}, \InpaintOQ{} and \InpaintSTOQ{}, where
    the subscripts indicate which of the above datasets were used.
We chose these subsets to study how the type and quantity of training data affect inpainter quality:
    \InpaintST{} only uses open-domain conversational datasets that do not contain any explicit question answering,
    while \InpaintOQ{} only uses relatively small conversational question answering datasets,
    and \InpaintSTOQ{} uses all of the above.
Each inpainter model was initialized from a pre-trained T5-XXL (11B parameters) checkpoint and finetuned on its corresponding training data.
See \refapp{appendix:inpainter-hparam} for details on the hyperparameters used.

\textbf{Document processing.}
We apply dialog inpainting to two document corpora:
\textsc{Wiki}, a collection of 11.4M passages from 5.9M English Wikipedia articles in the \orquac{} retrieval corpus~\citep{Qu2020ORQuAC}, and 
\textsc{Web}, a collection of 8.4M English web passages from the \msmarco{} retrieval corpus~\citep{Nguyen2016MSMARCO}.
Both corpora were used as is without any further filtering.
The passages were split into sentences using the Google Cloud Natural Language API.
To limit computation,
    we only use the first 6 sentences of each passage.\footnote{%
We truncated about 64\% of passages containing 7+ sentences.
}
The passages were then converted to partial dialogs and inpainted using the method described in \refsec{di_inference_details}.
The final results are the datasets \WikiDialogST{}, \WikiDialogOQ{} and \WikiDialogSTOQ{} and
\WebDialogPT{}, where the subscripts indicate which inpainter model was used to generate the data (\reftab{inpainter-data}).

\section{Evaluating \WikiDialog{} as a Dataset}
\label{sec:dataset}
We now turn to a qualitative evaluation of the dialogs generated in \refsec{method}.
We will show that they largely contain \emph{information-seeking} dialogs
    with well-matched questions and answers.
This makes the data suitable for ConvQA systems, an application explored later in the paper.
As \WikiDialog{} is generated using \emph{passages} from \orquac{},
    the corresponding \orquac{} \emph{dialogs} form a natural point of reference:
both datasets rely on the same content, but are respectively automatically and manually generated.  
For this reason, our analysis focuses on the \WikiDialog{} variants.


Our analysis combines automatic and human evaluation.
The human evaluation was conducted by asking human raters the subjective questions listed in \reftab{human-eval} for each turn of a dialog.\footnote{%
    Additional details, including screenshots of the annotation interface, instructions and examples, are provided in
    \refapp{appendix:dataset-analysis}.
}
We found substantial inter-annotator agreement on all four questions, with a Krippendorff's $\alpha$ of at least 0.89,    
and report results aggregated over dialogs corresponding to a common set of 200 randomly chosen passages using the mean score of three raters.

\begin{table}[t]
  \centering
  \newcommand{\ssig}[1]{\underline{#1}}

\begin{tabular}{lrrrr}
\toprule 
\textbf{\textit{Question}} ($\alpha$) & \multicolumn{3}{c}{\textbf{\WikiDialog{}}} & \textbf{OR-} \\ \cmidrule{2-4}
\quad \textbf{Answer}
    & \textbf{PT} & \textbf{OQ} & \textbf{PTOQ} & \textbf{QuAC} \\ \midrule
\multicolumn{5}{p{0.9\columnwidth}}{\textit{
    Is the question \emph{information seeking}?
     (0.98)}} \\
\quad Yes 
    & \ssig{94.5\%} & 100\% & 99.3\% & 100\% \\
\multicolumn{5}{p{0.9\columnwidth}}{\textit{
 How \emph{relevant} is question to the conversation?
  (0.94)}} \\
\quad Not at all &  0.3\% 	      & 0\%    & 0\% 		   & 0\% \\
\quad Topic only &  \ssig{45.8\%} & 49.5\% & \ssig{42.1\%} & 52.6\% \\
\quad Follows up &  \ssig{53.9\%} & 50.5\% & \ssig{57.9\%} & 47.4\% \\

\multicolumn{5}{p{0.9\columnwidth}}{
    \textit{How \emph{specific} is the question?
    (0.91)}} \\
\quad Not at all & 6         & \ssig{12}\%   &  5.8\% 		 & 5.4\% \\
\quad Somewhat   & 15\% 	 & \ssig{28.7\%} & \ssig{22.4\%} & 12\% \\
\quad Very       & 79\% 	 & \ssig{59.3\%} & \ssig{71.7\%} & 82.6\% \\
\multicolumn{5}{p{0.9\columnwidth}}{\textit{
    How \emph{well answered} is the question?
    (0.89)}} \\
\quad Not at all    &  0.1\%        & 0.1\%         &  0.0\%        & 0.8\% \\
\quad Incompletely  & \ssig{15.6\%} & 19.7\%        & 25.4\%        & 22.8\% \\
\quad Sufficiently  & \ssig{52.4\%} & \ssig{46.8\%} & \ssig{50.2\%} & 36.6\% \\
\quad Perfectly     & \ssig{31.9\%} & 33.4\%        & \ssig{24.4\%} & 39.8\% \\
\bottomrule
\end{tabular}

  \caption{\label{tab:human-eval}
  Results from a human evaluation of the generated utterances in three
  \WikiDialog{} variants vs. 
  a dataset manually collected on the same passages, \orquac{}.
  The table also presents inter-annotator agreement using Krippendorff's $\alpha$.
  Underlined numbers statistically differ from corresponding \orquac{} ones at a $p < 0.05$ level based on a paired randomization test.
  }
  \vspace{-1em}
\end{table}


\begin{figure}[t]
    \centering
    \includegraphics[width=0.9\columnwidth]{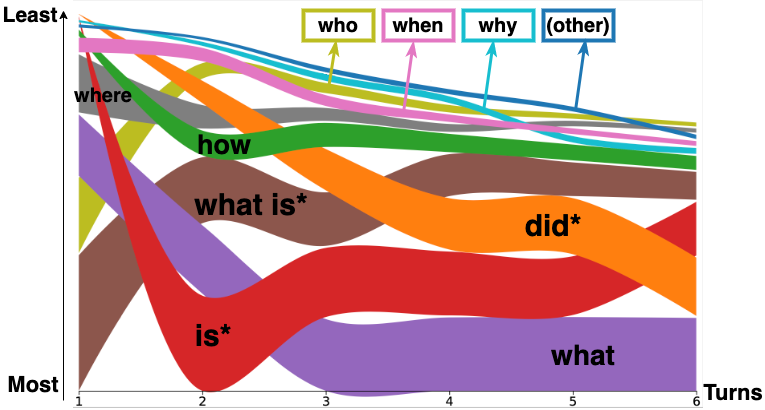}
    \caption{\label{fig:question-distribution}
    Bump chart showing how questions are distributed across turns in \WikiDialogSTOQ{}:
    the width and ordering of each line is proportional to the fraction of questions of that type in each turn. 
    Dialogs tend to start with definitional questions (what is, who, where, etc.) and diversify into
    a wide range of follow questions (what happened, did, is, etc). 
    }
\vspace{-1em}    
\end{figure}

\paragraph{How information seeking are the generated utterances?}
Raters judge the generated utterances to almost always be information seeking and topically relevant.
We note that, prompt aside, our method is not tailored to explicitly generate information seeking utterances in any way.
It is remarkable then that raters found 94.5\% of utterances in \WikiDialogST{}, 
    which was generated without using any ConvQA data,
    to be information seeking;
    by training the inpainter using ConvQA data, this number goes to 99--100\%.
Almost all of these are single sentences phrased as questions, though \WikiDialogST{} includes some exceptions, e.g.,
    \textit{``Exciting! I wonder if they got a \#1 debut with that song.''}.

\paragraph{What types of questions are generated?}
Next, we study the distribution of questions generated by clustering utterances on their first two words.
\reffig{question-distribution} shows how this distribution changes over turns:
    dialogs start with more definitional questions (e.g., what is, who is, where is, etc.) but then diversify into a range of follow-up questions (what happened, did, is, how, why, etc.).
A good conversation should have a mix of both specific and broad questions.
Here, raters find that the \WikiDialog{} variants primarily differ from \orquac{} in having more ``somewhat specific'' questions (e.g.,  \textit{``Is there anything else interesting about his guitar?''})
    and fewer ``very specific'' questions (e.g.,  \textit{``Did people enjoy their music?''}).

\paragraph{How well answered are the generated questions?}
So far, we have shown that one speaker in \WikiDialog{} consistently asks questions,
    playing the role of the imagined reader well.
We now turn to the other speaker, the writer:
    how often are their responses---sentences from a Wikipedia passage---adequate answers to the questions?
Raters find that questions in all three variants of \WikiDialog{} are sufficiently answered about as often as in \orquac{},\footnote{%
    When considering the net percentage of questions sufficiently or perfectly answered, the differences between the \WikiDialog{} variants and \orquac{} are \textit{not} statistically significant at the $p < 0.05$ level.
}
though questions in \orquac{} are often answered better.


\paragraph{How conversational are the data?}
A key distinguishing feature of dialog data is its dependence on prior dialog context.
    Following prior work~\citep{Choi2018QuAC}, we use
    the degree of context dependence as a measure of conversation.
Raters judge that questions in the \WikiDialog{} variants follow-up
    on dialog context significantly more often than those in \orquac{}.
We hypothesize that the tendency to follow-up likely arises from the structure of the 
underlying passages from which \WikiDialog{} is derived:
    when a paragraph in Wikipedia explores a single topic in depth, 
    so too does the dialog generated from it in \WikiDialog{}.
    


\paragraph{Do the data contain potentially sensitive language that may perpetuate unfair bias?}
Evaluating a dataset of this size for potential unfair bias is a daunting task that requires diligent and thorough investigation.
We approached the problem by curating a set of 700 terms from the literature~\citep{bolukbasi2016man,garg2018word,may2019measuring,nadeem2020stereoset,abid2021large} related to sensitive characteristics---such as race, ethnicity, gender, and sexual orientation.
Many instances of these terms are well-motivated: for example, a dialog from a passage about transgender rights in Canada includes the question ``What does anti-discrimination act mean in relation to \emph{transgender} people?''.
We further refined the approach to instead look at co-occurrences between these terms and adjectives that may have negative connotations, focusing on instances where the terms were not explicitly mentioned in the passage.
We find that 0.2--0.5\% of dialogs in the dataset contain such potentially sensitive interactions, but it is difficult to establish if they perpetuate unfair bias without expert manual review.
Therefore, we advise users to note these observations and exercise care while using the dataset.\footnote{%
Further details of our approach and some examples are provided in \refapp{appendix:safety}. 
}

\section{Application: Open-domain Conversational Retrieval}
\label{sec:application}

\begin{figure}[t]
  \centering
    \includegraphics[width=0.8\columnwidth]{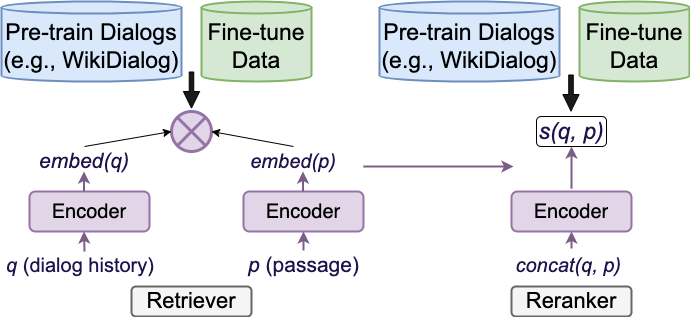}
\caption{\label{fig:retrieval-overview}
We use a two-stage ConvQA retrieval system. We first retrieve top-K passages from the corpus using a dual-encoder model and then rerank them using a cross-attention model.
}
\vspace{-1em}
\end{figure}

In this section, we show how our inpainted datasets can serve as a powerful source of training data for ConvQA systems.


A ConvQA system engages with a user through multi-turn dialog, where typically the user poses questions and the system answers (there can be exceptions, e.g., the system asks a clarifying question).
During a dialog, whenever it is the system's turn to speak (at some time $t$), it looks at all previous dialog turns $d_{1:t} = (u_1, u_2, \ldots, u_{t})$ which we call the \emph{dialog history}, and outputs a new utterance, $u_{t+1}$.

Because ConvQA dialogs are knowledge-intensive, many systems decompose the task into a two-part retrieve-then-generate process~\cite{Qu2020ORQuAC, Anantha2021QReCC}.
 First, they employ a \emph{conversational retriever} to retrieve passages that are relevant to the conversation based on the dialog history $d_{1:t}$. 
 Second, they employ a \emph{generator} which uses both the dialog history ($d_{1:t}$) and the retrieved passages to generate a response, $u_{t+1}$. While both steps are important, the \emph{conversational retriever} is key to helping the model access the right knowledge and also for showing people evidence for an answer. 
 
This work focuses on the conversational retriever, showing how to improve it by pre-training on our inpainted data, leaving improvements to the generator for future work.
\textbf{Models.}
The input to a conversational retriever is the dialog history ($d_{1:t}$) and a passage ($p$). The output is a score, $s(d_{1:t}, p)$, indicating the passage's relevance. Retrieval is performed by selecting the passages with the highest scores.

We also refer to the dialog history as the ``query'' and denote it as $q$. In some benchmarks that we study, the ``dialog history'' is defined to be all previous utterances, while in others the history is defined to only include the user's questions but not the system's responses. 

We employ two standard models for retrieval: first, we use a \emph{dual encoder}~\cite{Reimers2019SentenceBERT, Karpukhin2020DPR, Ni2021SentenceT5} to select an initial set of candidates. We then rescore those candidates using a cross-attention \emph{reranker}~\cite{Nogueira2019BertReranker, Nogueira2020T5Reranker}. Model architectures and training objectives are detailed in \refapp{appendix:retrieval_models}. To train these models, we need a corpus of $(q, p^*)$ pairs, where $q$ is a dialog history and $p^*$ is a relevant passage. The following sections describe such data. Figure~\ref{fig:retrieval-overview} illustrates our system. 

\textbf{Pre-training and Fine-tuning.}
As outlined in Section~\ref{sec:di_inference_details}, each dialog generated by our inpainter tends to consist of alternating question and answer utterances:
$d = (s_\text{prompt}, \hat{u}_1, s_1, ..., \hat{u}_m, s_m)$,
where inpainted utterances $\hat{u}_i$ are questions, 
and their subsequent answers $s_i$ are sentences from the original passage $p$.
Intuitively, for each question in the dialog, $p$ is a highly relevant passage that should be retrieved.
Based on this observation, we generate examples as follows: first, we randomly select a dialog prefix that ends in a question to be the dialog history: $q_i = (\hat{u}_1, s_1, \ldots, \hat{u}_i$).\footnote{We omit the leading prompt utterance ($s_\text{prompt}$) from the history. Also, we optionally omit answer turns, depending on the task setup.} We then wish to mark the original passage $p$ as a positive passage to retrieve. However, directly using $p$ as a positive example will not yield good results: the dialog history ($q_i$) includes exact sentences from $p$, which would cause our retriever to simply learn to string-match, rather than to generalize.
To eliminate this problem, we form a new passage that consists only of the remaining sentences in $p$ that haven't appeared in $q_i$ yet:
$p^*_i \eqdef \Concat({s_j\text{ where } j > i})$.

After pre-training $(q_i, p^*_i)$ pairs from the inpainted data, our retriever is fine-tuned on a downstream ConvQA dataset.
\section{Evaluation}
\label{sec:evaluation}

We report quantitative evaluation of dialog inpainting by measuring the impact of \WikiDialog{} and \WebDialog{} on ConvQA retrieval systems.

\subsection{Experimental setup}
\label{sec:eval-setup}
The following presents a summary of our experimental setup;
see \refapp{appendix:evaluation} for full details.

\begin{table}[t]
  \centering

\begin{tabular}{l rr r}
\toprule
\textbf{Dataset} & \textbf{Dialogs} & \textbf{Turns} & \textbf{Ques.} \\
\midrule 
\multicolumn{4}{l}{\textit{Non-conversational question answering datasets}} \\
\msmarco{}          &       &           & 0.5M   \\
PAQ        &       &           &  65M   \\
\multicolumn{4}{l}{\textit{Non-question answering conversational datasets}} \\
\socialmedia       & 2.0M  &  1:2:4    &        \\
\taskmaster        & 19.8K & 9:10:14   &        \\
\multicolumn{4}{l}{\textit{Conversational question answering datasets}} \\
\orquac            &  5.6K & 4:8:12    & 40.5K \\
\qrecc             & 13.6K & 1:6:12    & 80.0K \\
CAsT-19    &    80 & 7:10:12   &   748 \\
CAsT-20    &    25 & 6:8:13    &   216 \\
\midrule 
\WikiDialog{}       & 11.4M & 1:6:6     & 56.1M \\
\WebDialog{}        & 8.4M & 1:3:6      & 26.9M \\
\bottomrule
\end{tabular}

  \caption{\label{tab:inpainter-data}
  The number of dialogs, turns and  questions (Ques.)
    of datasets considered in this paper;
  turns and question lengths are reported using (1\%:50\%:99\%) percentiles.
  }
  \vspace{-1em}
\end{table}

\textbf{Datasets.}
We use three open-domain conversational QA retrieval benchmarks:
    \orquac{}~\citep{Qu2020ORQuAC}, \qrecc{}~\citep{Anantha2021QReCC}, and     TREC \cast19{} and \cast20{}~\citep{Dalton2019TRECCAst19, Dalton2020TRECCAst20}. 
\reftab{inpainter-data} summarizes basic statistics of the each dataset; further details are provided in \refapp{appendix:baselines}.

\textbf{Retrieval corpora.}
We use standard retrieval corpora and relevance labels from each dataset for evaluation evaluation: 
\orquac{} uses 11.1M passages from English Wikipedia.
TREC \cast-19 and \cast-20 use a shared corpus of 8M web passages from \msmarco{}~\citep{Nguyen2016MSMARCO} and 30M Wikipedia passages from \citet{Dusart2019TRECCAR};
\qrecc{} uses 54M passages from the Internet Archive Wayback Machine and the English pages in CommonCrawl.
Following prior work~\citep{Qu2020ORQuAC, Yu2021ConvDR, Wu2021CONQRR}, we use only the current and previous
questions as inputs for OR-QuAC, TREC CAsT-19 and CAsT-20, and use previous answers in addition to these as input for QReCC.
We report mean reciprocal rank at rank 5 (MRR@5) for \orquac{} and  mean reciprocal rank at full rank (MRR) for \qrecc{} and TREC CAsT. Additional metrics are reported in \refapp{appendix:results}. 




\textbf{Model implementation.} 
We initialize our dual-encoder retrievers and rerankers from pre-trained T5 checkpoints following prior work~\cite{Ni2021SentenceT5, Nogueira2020T5Reranker}.
Both retrievers and rerankers are pre-trained on our inpainted datasets. 
    We use the notation T5-Large DE $\triangleright$ WikiD to denote a dual-encoder that was
    initialized from a T5-Large checkpoint and pre-trained on \TextWikiDialog{} (we use \TextWikiDialogSTOQ{} unless otherwise specified), and T5-Large DE $\triangleright$ WikiD+WebD to denote a model pre-trained on the union of \TextWikiDialog{} and \TextWebDialog{}.
Similar notation is used for T5-Base initialized models and rerankers. 
    
    
During fine-tuning,
    we separately train retrievers and rerankers on \orquac{} and \qrecc{}, using their validation sets to select checkpoints.
Because \cast19{} and \cast20{} are extremely small datasets and do not include a training split,
    we  do not fine-tune \emph{dual-enocoder retrievers} on these datasets, instead using a retriever finetuned on \qrecc{} data\footnote{The QReCC training data does not include TREC CAsT.}. 
We follow~\citet{Yu2021ConvDR}
    and use 5-fold cross-validation to finetune \emph{rerankers} on \cast19{} and \cast20{}: for each fold, we split the data into 5 splits based on dialogs, train a reranker on 3 splits of the data, select a checkpoint on one split and test on the remaining split. 
Similar to \citet{Yu2021ConvDR} and \citet{lin2021TCT}, we use hard negatives mined from a base retriever during fine-tuning (See \refapp{appendix:de-multi-stage-training} for details).
Finally, during inference, we use MIPS~\cite{Ram2012MIPS} to index the passage embeddings, and use exhaustive search to retrieve and then rerank the top 100 passages for all datasets.

\textbf{Baselines.}
We compare our pre-trained dual-encoder retrievers against five published \emph{retrieval-only} baselines:
BM25-Query Rewriter~\cite{Yu2020FewShotQueryRewrite},
BM25-T5QR~\cite{Wu2021CONQRR},
ANCE-Query Rewriter~\cite{Yu2021ConvDR},
CONQRR~\cite{Wu2021CONQRR}
and
ConvDR~\cite{Yu2021ConvDR}.\footnote{\refapp{appendix:baselines} includes brief summaries of these methods.}
We also evaluate our base model, a T5-Large DE model that is only finetuned on each dataset:
    this is a strong baseline that outperforms previous state-of-the-art on \qrecc{}. 
We also compare our pre-trained \emph{retrieval+reranking} systems against the best performing systems in TREC CAST\footnote{TREC CAsT has several settings based on if a model uses additional inputs such as manual query rewrites. We follow the \emph{automatic} setting that does not use additional inputs, and compare against top \emph{automatic} runs reported in ~\citet{Dalton2019TRECCAst19, Dalton2020TRECCAst20}.}:
CFDA\_CLIP\_RUN7~\cite{Yang2019CFDA} and h2oloo\_RUN4~\cite{Dalton2020TRECCAst20}, and a state-of-the-art retriever+reranker method
ConvDR$\rightarrow$BERT~\cite{Yu2021ConvDR}. 

\subsection{Main Results}
\label{sec:eval-main}


\begin{table}[t]
\centering
\setlength{\tabcolsep}{5pt}
\small
\begin{tabular}{l c c c c}

\toprule
 & & & \multicolumn{2}{c}{TREC CAsT} \\
            & \multicolumn{1}{c}{\qrecc{}} & \multicolumn{1}{c}{\orquac{}} & \multicolumn{1}{c}{19} & \multicolumn{1}{c}{20} \\ 
            \cmidrule(lr){2-2}   \cmidrule(lr){3-3}      \cmidrule(lr){4-5}  
\textbf{System} 
    & \small MRR   
    & \small MRR@5
    & \small MRR  
    & \small MRR  
    \\ \midrule 
& \multicolumn{4}{c}{\textit{Retrieval}}                         \\
BM25-QR     & --    & 20.2  & 58.1  & 25.0    \\
ANCE-QR     & --    & 45.7  & 66.5  & 37.5    \\
ConvDR      & --    & 61.6  & 74.0  & 50.1    \\
BM25-T5QR   & 32.8  & --   & --    & --      \\
CONQRR      & 41.8  & --   & --     & --      \\  \cmidrule(l){1-1}
T5-Large DE & 55.7  & 56.9 & 61.0 & 34.3\\ 
\quad $\triangleright$ \small{WikiD}
            & 60.4  & 66.5  & 68.1  & 43.7  \\   
\quad $\triangleright$ \textbf{\small{WikiD+WebD}} 
            & \textbf{60.7} & \textbf{68.7}  & \textbf{74.1}  & \textbf{51.3}   \\ \midrule \midrule
& \multicolumn{4}{c}{\textit{Retrieval + Reranking}}                  \\ 
CFDA\_CLIP\_RUN7 
            & --    & --    & 71.4   & -- \\
h2oloo\_RUN4
            & --    & --   &  -- & 59.3   \\
ConvDR$\rightarrow$BERT 
            & --     & 77.3  & 79.9  & 54.5   \\ \cmidrule(l){1-1}
T5-Large DE   & \multicolumn{3}{c}{} \\
$\rightarrow$ reranker
            & 68.9   & 72.6  & 75.3 & 55.1 \\
\quad $\triangleright$ \small{WikiD} 
            & 70.7  & 79.7&  79.3 & \textbf{60.3} \\
\quad $\triangleright$ \textbf{\small{WikiD+WebD}} 
            & \textbf{71.8} & \textbf{81.2} & \textbf{82.0} & 59.7 \\
\bottomrule
\end{tabular}
\caption{\label{tab:main-results} We evaluate  models trained using our inpainted datasets against baselines on three different ConvQA retrieval tasks: \qrecc{}, \orquac{}, and TREC \cast{}. WikiD, WikiD+WebD: We use \WikiDialogSTOQ{} for all tasks except when WikiD+WebD is used in TREC CAsT, where \WikiDialogST{} gives higher performance.  Additional metrics are reported in \refapp{appendix:results}.
}
\end{table}

\reftab{main-results} compares our models with baselines on four ConvQA retrieval benchmarks. 
We first note that our relatively simple base retriever model, T5-Large DE, is a strong baseline
    and significantly outperforms the existing state-of-the-art results on \qrecc{}.
By simply pre-training on \WikiDialog{}, we observe a 9--30\% gain, and outperform all baselines on  \qrecc{} and \orquac{} with large margins. Including \WebDialog{} increases this gain by a further 3--15\% MRR, with the most significant gains accruing to \cast-19 and \cast-20.
With this model, we observe  a 43\% relative MRR gain over the current state-of-the-art on \qrecc{}, a 12\% relative MRR@5 gain on \orquac{}, and comparable or better performance on \cast-19 and \cast-20. 
    

The gains achieved in the retrieval stage remain when the results are reranked: 
we outperformed the current state-of-the-art on \orquac{} and TREC \cast{}-19, and achieved comparable performance on \cast{}-20.
We could not find existing published reranker results for \qrecc{}, but observe a 5\% relative MRR gain over reranking the non pre-trained T5-Large DE model, which we found to be a strong baseline.  

%

\subsection{Analysis}
\label{sec:eval-analysis}
Next, we dive into our results in greater detail and study factors contributing to performance:

\begin{table}[t]
\centering

\small
\begin{tabular}{l c c c}
\toprule
        & \multicolumn{1}{c}{\qrecc{}} & \multicolumn{1}{c}{\orquac{}} & \multicolumn{1}{c}{CAsT-19}  \\
           \cmidrule(lr){2-2} \cmidrule(lr){3-3} \cmidrule(lr){4-4}
System & MRR  & MRR@5  & MRR \\ \midrule
T5-Base DE &  53.4 & 53.6   & 55.2   \\                                                         
~~$\triangleright$ \small{\msmarco{}}
          & 55.0 & 57.5   & 57.5          \\ 
~~$\triangleright$ \small{PAQ}
          & 57.0 & 62.7  & 48.4   \\
~~$\triangleright$ \small{PT}
        & 56.8 & 59.5  & 60.1  \\ \cmidrule(l){2-4}
~~$\triangleright$ \small{$\textrm{WikiD}_{\textrm{PT}}$}
         & 58.5 & 64.0   &   61.3     \\ 
~~$\triangleright$ \small{$\textrm{WikiD}_{\textrm{OQ}}$}
         & 58.9 & \textbf{65.3}  & 65.5  \\
~~$\triangleright$ \small{$\textrm{WikiD}_{\textrm{PTOQ}}$}
        & \textbf{59.0} & 64.1  & \textbf{66.5}  \\
\bottomrule
\end{tabular}

\caption{\label{tab:ablation-by-de-pretrain}
Retriever performance when T5-Base DE is pre-trained ($\triangleright$) on various datasets, then fine-tuned on \qrecc{} and \orquac{}; results on \cast-19 use a retriever finetuned on \qrecc{}. 
}
\vspace{-1em}
\end{table}

\paragraph{How much does in-domain training matter for inpainter quality?}
\reftab{ablation-by-de-pretrain} compares the performance of retrievers
pre-trained on three variants of \TextWikiDialog{} introduced in \refsec{dataset}: PT, OQ, and PTOQ.
Remarkably, using \TextWikiDialogST{} is sufficient to significantly outperform current state-of-the-art methods,
    despite being generated using only open-domain conversational data.
Next, we observe that using an inpainter trained on \orquac{} and \qrecc{} data (\TextWikiDialogOQ{})
    results in slightly better performance;
    it is notable that an inpainter trained using only ~20K dialogs is able to generate such high-quality data, and
    can improve performance over a retriever just fine-tuned on the same data by up to 20\%.
Finally, we observed that an inpainter trained on both types of data (\TextWikiDialogSTOQ{}) is able to generalize
    better to \cast-19, an out-of-domain task we evaluate on using a \qrecc{} retriever.


\paragraph{How does \TextWikiDialog{} compare to other retriever pre-training datasets?}
\reftab{ablation-by-de-pretrain} also reports results for 
retrievers pre-trained on several alternatives to \TextWikiDialog{}:
\msmarco{}~\cite{Nguyen2016MSMARCO}, a search dataset that is widely used for retriever pre-training;
\paq{}~\cite{Lewis2021PAQ}, the largest existing (non-conversational) question-answering dataset,  also automatically generated from Wikipedia;
and \socialmedia{} + \taskmaster{} (PT), the open-domain dialog data introduced in \refsec{di_task} as a training dataset for the inpainter---we use it here to directly pre-train a conversational retriever using a next utterance retrieval task.  \reftab{inpainter-data} lists their characteristics.

We observe that pre-training on any of the \WikiDialog{} variants significantly outperforms the two classic non-conversational QA datasets, \msmarco{} and \paq{}. \WikiDialog{} also outperforms open-domain dialog data \socialmedia{}+\taskmaster{} (PT).  Despite being generated using PT, pre-training on \TextWikiDialogST{} far outperforms it,
    showing that the proposed dialog inpainting recipe provides additional signal not present in PT. 

In addition, we found that  
pre-training a retriever on PT does surprisingly well, significantly outperforming \msmarco{} on all datasets. 
Prior ConvQA retrieval systems have focused on traditional QA  data for pre-training, but we find that open-domain conversational data can be just as useful!

\begin{figure}[t]
  \centering
  \includegraphics[width=\columnwidth]{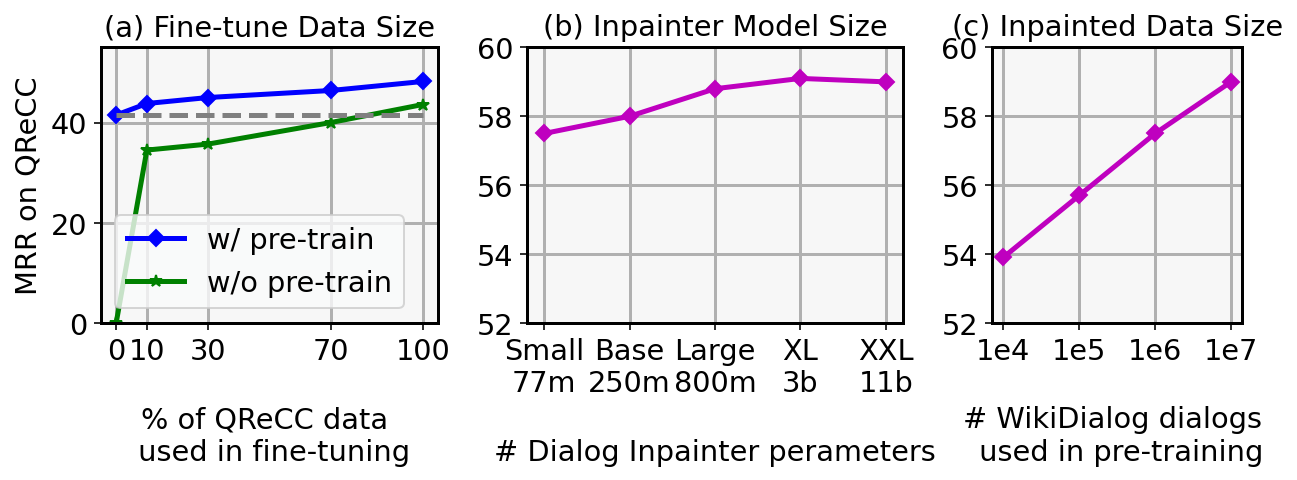}
\caption{\label{fig:ablations} Retriever performance on QReCC when T5-Base DE $\triangleright$ \WikiDialogST{} is trained with (a) varying fine-tuning data sizes, (b) different sizes inpainter models, and (c) varying pre-training data sizes.
Results in (a) do not include mined hard-negatives.
}
\vspace{-1em}
\end{figure}

\paragraph{Does pre-training on WikiDialog enable zero/few-shot learning?}
We now explore how much fine-tuning data is needed after pre-training on \TextWikiDialog{}.
\reffig{ablations} (a) plots the retrieval results of a T5-Base retriever pre-trained on \TextWikiDialogST{}---which does not use any in-domain data---and finetuned on varying percentages of the \qrecc{} training dataset \footnote{Unlike the other experiments presented in this paper that uses mined hard negatives to fine-tune retrievers,   
    these results do not use hard negatives because our hard negative mining model was fine-tuned on the whole \qrecc{} dataset (\refapp{appendix:retrieval_models}).}. 

Surprisingly,
    we observe that the zero-shot performance of the pre-trained retriever is already quite good: 
    achieving nearly 95\% the MRR of a model that uses the full dataset.
Furthermore, pre-trained retriever needs fewer fine-tuning data -- by fine-tuning on just 10\% of the data,
    the pre-trained retriever exceeds the performance of the full-data baseline.

\paragraph{Does our method scale with inpainting model size and data size?}
We now explore if our dialog inpainting method can benefit from scaling up along two dimensions: the inpainter model size, and the inpainted \WikiDialog{} data size. Results are shown in \reffig{ablations} (b) and (c).

From \reffig{ablations} (b), we observe that retriever performance increases with inpainter model size with one exception: 
the T5-XL  model slightly outperforms T5-XXL;
we hypothesize this is due to insufficient hyperparameter search for T5-XXL. Surprisingly, the quality of data generated by T5-Small is already 
sufficient to significantly outperform current state-of-the-art methods.

In \reffig{ablations} (c), we evaluate how retrievers pre-trained with 10K--11M dialogs sampled from \TextWikiDialog{}
perform on \qrecc{}.
We observe a roughly log-linear relationship between performance and pre-training data size that
has not yet plateaued:
    simply inpainting more passages may further increase retrieval performance. 

\section{Related Work}
\label{sec:related-work}

\textbf{Conversational question answering retrieval.}
Several manually collected conversational question answering datasets have been proposed to address the 
    scarcity of high-quality training data
~\citep{Choi2018QuAC,reddy2018coqa,Dinan2018wow,Saeidi2018sharc,Dalton2019TRECCAst19,Campos2020doqa,Dalton2020TRECCAst20,Qu2020ORQuAC,Feng2020doc2dial,Anantha2021QReCC}.
However, because they are relatively small,
existing retrieval systems all depend on rewriting queries to use with a \emph{non-conversational} retrieval system~\citep{Yang2019CFDA,Dalton2020TRECCAst20,Yu2020FewShotQueryRewrite,Yu2021ConvDR, Wu2021CONQRR}.
Query rewriting is a hard problem in itself~\citep{vakulenko2020wrong}---prior systems rely on proprietary search logs~\citep{Yu2020FewShotQueryRewrite}, reinforcement learning~\citep{Wu2021CONQRR} or distillation recipes~\citep{Yu2021ConvDR}---and some queries cannot be rewritten at all, e.g., ``What \emph{else} were they famous for?''.
Here, we use \WikiDialog{}, a 1000x larger ConvQA dataset, to train a standard retriever without requiring query rewrites.

\textbf{Conversational language models.}
Large conversational language models such as DialogGPT~\citep{ZhangSGCBGGLD20} and Meena~\citep{adiwardana2020towards}
have shown impressive open-ended conversational capabilities, and even the ability to directly answer many questions.
However, they often hallucinate answers and amplify unfair biases present in their training data.
Subsequent work address this problem by instead posing queries to a \emph{non-conversational} retriever and using its answers~\citep{Roller2020blender1, Komeili2021Blender2,nakano2021webgpt,thoppilan2022lamda, Elgohary2019CANARD}. 
In this paper, we instead use a \emph{masked} conversational language model to generate ConvQA data:
    the data can be readily audited for unfair biases 
    and to train a \emph{conversational} retrieval system. 

\textbf{Data augmentation and synthetic data generation.}
Data augmentation has been widely used to improve the performance of document retrieval and related tasks~\citep{LeeCT19,ChangYCYK20,Gao2021SimCSE} by using training data more efficiently.  
However, these methods typically do not result in high-quality data that can be used outside of a training recipe.
More recently, \citet{MaKYHM21} and \citet{Lewis2021PAQ} train special-purpose question generation models 
    to synthetically generate millions of high-quality question answer pairs.
We extend this line of work to generate millions \emph{information seeking dialogs},
    and show that even a general purpose inpainter model can generate high-quality data.

\section{Discussion}
\label{sec:discussion}

In this paper, we have presented \emph{dialog inpainting},
a novel approach to generating synthetic \emph{conversational} data.
We showed that it is possible to generate compelling information-seeking dialogs using only general-purpose data, suggesting applications to other conversational tasks.
While synthetic data cannot entirely replace real data, it can help bootstrap interactive conversation systems and create a virtuous cycle wherein users find it natural to engage with and improve the system.
We are particularly optimistic about applying the dialog inpainting data to (1) distillation, where the inpainted datasets serve as large-scale distillation sets, (2) end-to-end conversational question answering, and (3) \emph{zero-shot} conversational QA, which is motivated by the zero-shot retrieval capabilities shown in this work.

It is important to be aware of the biases that generating data can introduce or amplify.
We want to encourage good inductive biases that make conversations conversational---e.g., use of anaphora or elision of context---and to introduce further control over the dialogs generated---e.g., persona or dialog acts.
At the same time, we must interrogate the generated data and work towards minimizing instances of potentially sensitive language that may perpetuate unfair biases. 


\bibliography{icml2021}
\bibliographystyle{icml2022}

\appendix
\section{Inpainting Models}
\subsection{Datasets for \inpainter{} training}
\label{sec:appendix:train_dataset}

\begin{enumerate}
\item 
    \socialmediab{} is an open-domain dialog dataset proposed by LaMDA~\citep{thoppilan2022lamda}.
    Each dialog is mined from public forums and then scored and filtered using LaMDA's SSI and safety model.
    While the original dataset contains multi-speaker dialogs, we only use a subset consisting of two-speaker dialogs.
    Note that most of the dialogs in PublicDialog are short and have only two turns.
\item 
    \taskmasterb{}\footnote{%
        There are three datasets, Taskmaster-1, Taskmaster-2, and Taskmaster-3; we use Taskmaster-1. Also see \url{https://github.com/google-research-datasets/Taskmaster}. 
    }~\cite{ByrneKSNGDYDKC19}
        is a crowd-sourced dataset of task-oriented dialogs between two speakers.
    We only use the subset of dialogs related to movie ticket and restaurant reservations.
    While \taskmaster{} is significantly smaller than \socialmedia{}, its dialogs have many more turns, which helps mitigate turn bias.
\item 
    \orquacb{}~\cite{Qu2020ORQuAC} is a version of the conversational question answering dataset QuAC~\cite{Choi2018QuAC} 
        that consists of real human-human QA dialogs about a given Wikipedia article. 
    \orquacb{} extends QuAC to the open-domain retrieval setting by making the first question context-independent and pairing each answer with a relevant passage from Wikipedia.
\item 
    \qreccb{}~\cite{Anantha2021QReCC} is another conversational question answering dataset constructed using question sequences from QuAC, TREC Conversational Assistant Track, and NaturalQuestions \cite{Choi2018QuAC,Dalton2019TRECCAst19,Kwiatkowski2019NQ}.\footnote{%
        The NaturalQuestions dataset does not originally contain question sequences, so the authors asked human annotators to come up with follow-up questions first.
    }
    For each question sequence, annotators search a document corpus to find relevant passages and answers.
    Though the questions in QReCC overlap with those in \orquacb{}, their answers can be completely different.
\end{enumerate}

\subsection{Training details}
\label{sec:appendix:inpainter-hparam}
Unless otherwise specified, all our \inpainter{}s are initialized from T5-XXL (11B parameters)\footnote{We use t5.1.1 checkpoints from \url{https://github.com/google-research/text-to-text-transfer-transformer/blob/main/released_checkpoints.md}} and finetuned using 64 TPU v3 chips \footnote{\url{ https://cloud.google.com/tpu/}} with constant learning rate 0.01, dropout rate 0.1 and batch size 128.
We trained the $\Inpaint_\textrm{PT}$ and $\Inpaint_\textrm{PTOQ}$ using 100k steps, and $\Inpaint_\textrm{OQ}$ using 10k steps because its training set is significantly smaller.

\subsection{Inference details}
\label{sec:appendix:inpainter-inference}
To generate the inpainted datasets, we used \href{Apache Beam}{https://beam.apache.org} to parallelize our computation.
On average, it took from ~20ms (for T5-Small) to -~141ms (for T5-XXL) to inpaint each utterance in a dialog,
and required between 100 TPUv3-hours (for T5-Small) and 1900 TPUv3-hours (for T5-XXL) to inpaint each \WikiDialog{} variant,
and would cost between \$240 and \$4560 using preemptible TPUs on Google Cloud.
\section{Human Evaluation Protocol}
\label{sec:appendix:dataset-analysis}
In this section we overview the human evaluation protocol used to evaluate \WikiDialog{} variants and \orquac{} in \refsec{dataset}; the results of this evaluation were presented in \reftab{human-eval}.

\begin{figure*}
\centering
\includegraphics[width=0.9\textwidth]{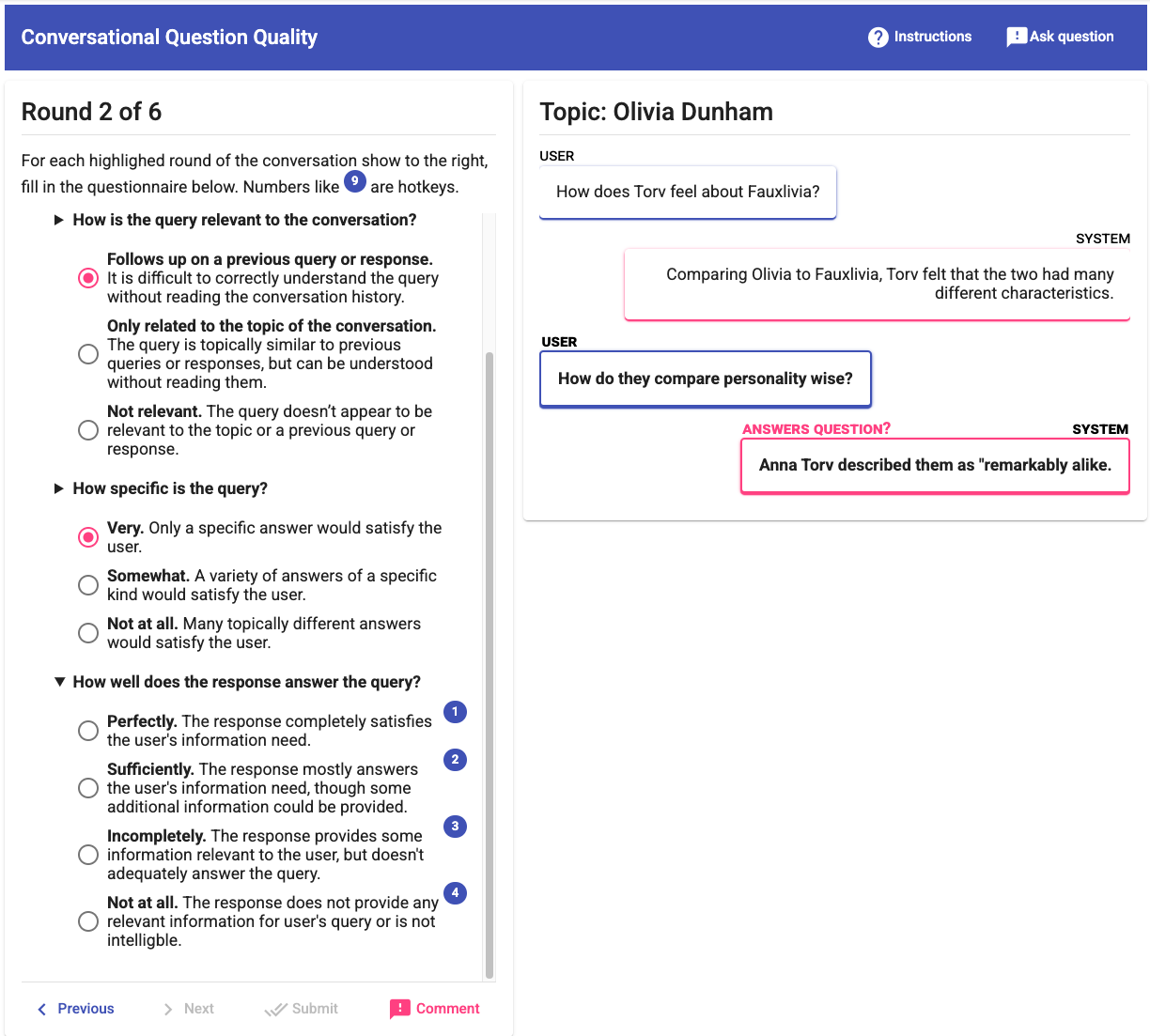}
\caption{\label{fig:annotation-interface} A screenshot of the annotation interface.
On the right, raters see the conversation history and the highlight turn they must rate in the form on the left.
When all questions for a given turn are completed, they are allowed to move forward
to the next turn (round in the figure).
The task can be submitted when all turns are complete. 
}
\end{figure*}

\begin{figure*}
\centering
\includegraphics[width=0.9\textwidth]{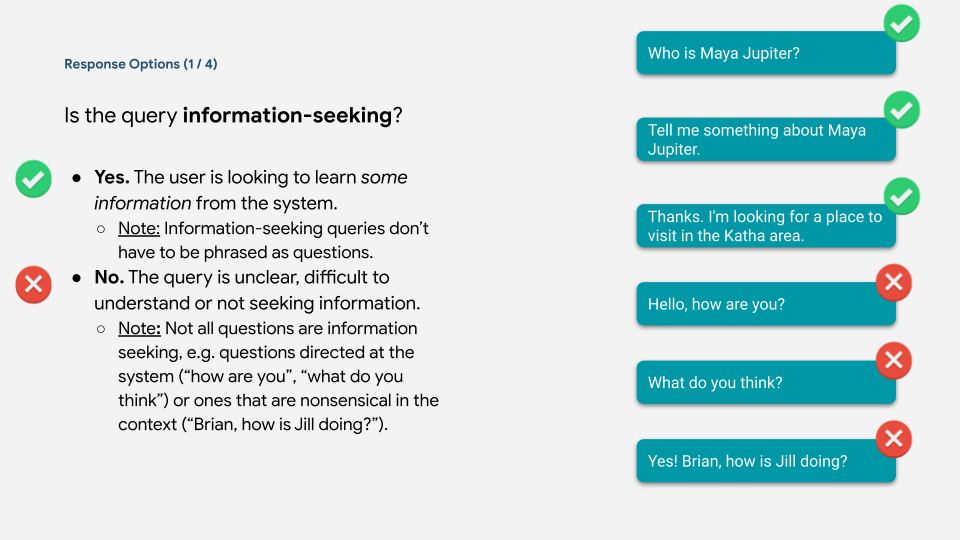}
\caption{\label{fig:instructions-q1} Instructions and example provided for the question ``Is the query information-seeking?''}
\end{figure*}

\begin{figure*}
\centering
\includegraphics[width=0.9\textwidth]{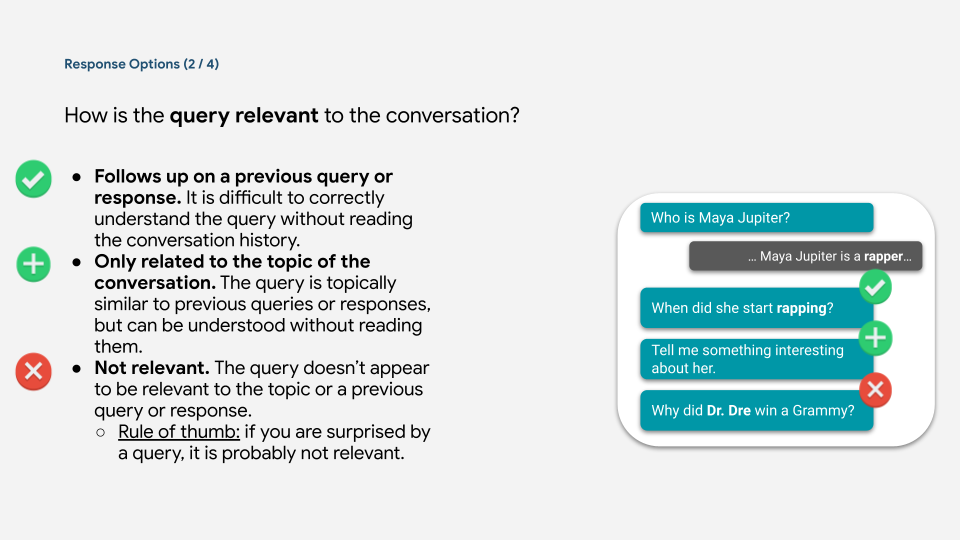}
\caption{\label{fig:instructions-q2} Instructions and example provided for the question ``How is the query relevant to the conversation?''}
\end{figure*}

\begin{figure*}
\centering
\includegraphics[width=0.9\textwidth]{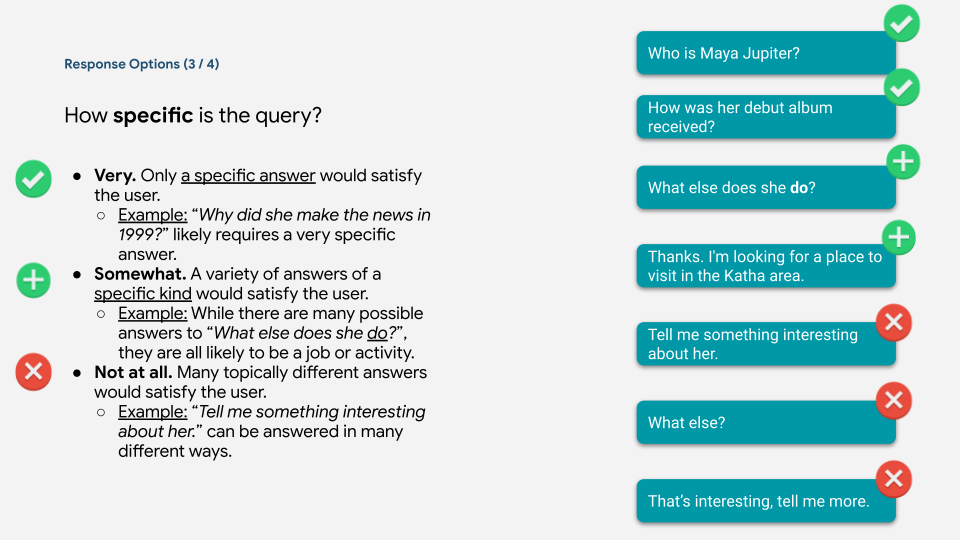}
\caption{\label{fig:instructions-q3} Instructions and example provided for the question ``How specific is the query?''}
\end{figure*}

\begin{figure*}
\centering
\includegraphics[width=0.9\textwidth]{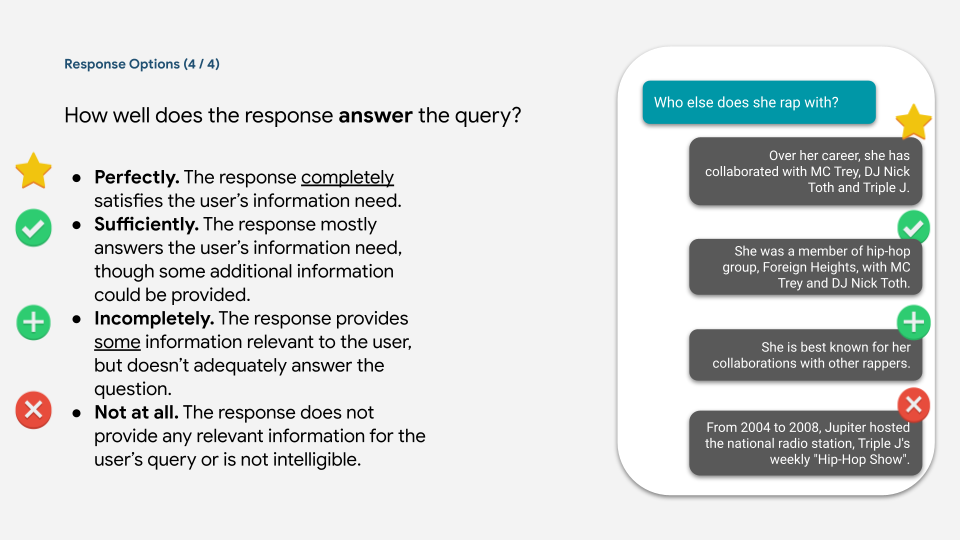}
\caption{\label{fig:instructions-q4} Instructions and example provided for the question ``How well does the response answer the query?''}
\end{figure*}

\begin{figure*}
\centering
\includegraphics[width=0.9\textwidth]{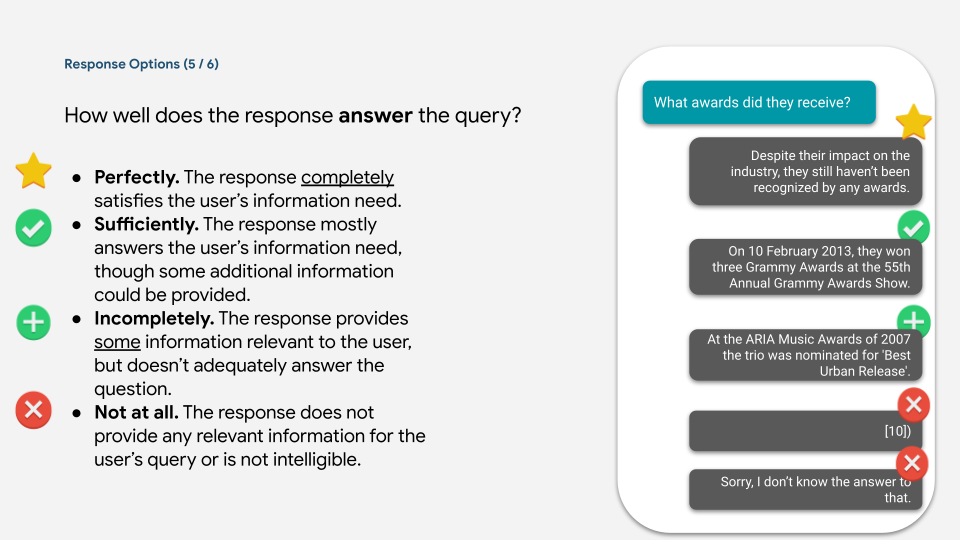}
\caption{\label{fig:instructions-q4b} An additional example for the question ``How well does the response answer the query?''}
\end{figure*}

\paragraph{Task design and iteration.}
\reffig{annotation-interface} provides a screenshot of the annotation interface and a description of its features.
The task was established as follows:
\begin{quote}
In this task, you will be spotting nonsensical or factually incorrect messages in artificially generated information-seeking conversations between a user and a system. 
\end{quote}
In each turn, raters were asked to answer the subjective questions in \reftab{human-eval} and were provided the instructions and adjoining examples in 
Figures~\ref{fig:instructions-q1}-\ref{fig:instructions-q4b}.

We iterated on our task design and instructions over two pilot annotations runs.
Feedback from these pilots led to two main changes:
(i) we extended the specificity and answer adequacy questions to be Likert scales rather than yes/no questions and
(ii) we omitted a laborious question that asked raters to attest whether the passage supports the answer when interpreted in the context of the conversation history---we found that this perfectly correlated with whether or not the question was adequately answered or not.

\paragraph{Rater recruitment and training.}
We engaged with a vendor supplier of full-time crowd workers to recruit human annotators for our task. Raters were asked to review the above instructions and were provided direct feedback on their responses during the pilot annotation runs.





\section{Retrieval models}
\label{sec:appendix:retrieval_models}

\subsection{Dual encoder retriever}
The dual-encoder maps a query ($q$) and a passage ($p$) into dense embedding vectors $\embed_\gamma(q)$ and $\embed_\gamma(p)$, where $\gamma$ denotes model parameters. The relevance score between the two is their vector cosine similarity:
\begin{equation*}
s_\gamma(q, p) = \frac{\embed_\gamma(q)^{\top}\embed_\gamma(p)}{ \|\embed_\gamma(q)\| \cdot \|\embed_\gamma(p)\|}
\end{equation*}
This particular function enables one to retrieve the top-K highest-scoring passages for a given query using fast similarity search methods that run in sub-linear time~\cite{Ram2012MIPS}.

For training, we minimize a standard contrastive loss with temperature $\tau$:
\begin{equation}
  \mathcal{L}(\gamma) = -log \frac{\exp (s_\gamma(q, p^*)/\tau)}{ \sum_{p \in p^* \cup \mathcal{N}(q)} \exp({s_\gamma(q, p)/\tau)}},
\end{equation}
where $p^*$ is a positive passage for $q$ and $\mathcal{N}(q)$ denotes negative passages.

We implement the dual-encoder retriever following recent work~\cite{Ni2021SentenceT5}: in particular, we use a shared Transformer encoder initialized from a T5 checkpoint, take the mean pooling of the top-most encoder layer, and project it to a fixed 768-dimensional embedding. 

\subsection{Reranker}
The reranker model takes the same inputs as the dual encoder, but instead of encoding $q$ and $p$ into two separate vectors, it jointly encodes them into a single vector, $\embed_\psi(q, p)$, where $\psi$ denotes model parameters. It outputs a relevance score:
\begin{equation*}
s_\psi(q, p) = w^{\top} \embed_\psi(q, p)
\end{equation*}
where $w$ is also a model parameter. Unlike the dual encoder, this function does not support fast top-K retrieval in sub-linear time. Therefore, for computational tractability, we only use it to rerank a short-list of candidates retrieved by the dual encoder. However, the joint embedding of $q$ and $p$ permits a more expressive relevance function (e.g. cross-attention between $q$ and $p$), so it can improve over the dual encoder's relevance scores.

For training, we minimize a weighted binary classification loss:
\begin{align*}
&\mathcal{L}(\psi) = \\
& -\log \sigma(s_\psi(q, p^*))
- \mathcal{N}(q)^{-1} \sum_{p \in \mathcal{N}(q)} \log [1 - \sigma(s_\psi(q, p))],
\end{align*}
where $\sigma$ denotes the sigmoid function. 

Similar to our retriever, the reranker is also initialized from a T5 encoder. Our reranker implementation follows the implementation described in ~\citet{Nogueira2020T5Reranker}.

\subsection{Model Training}
\label{sec:appendix:de-multi-stage-training}

\begin{figure}[t]
  \centering
    \includegraphics[width=\columnwidth]{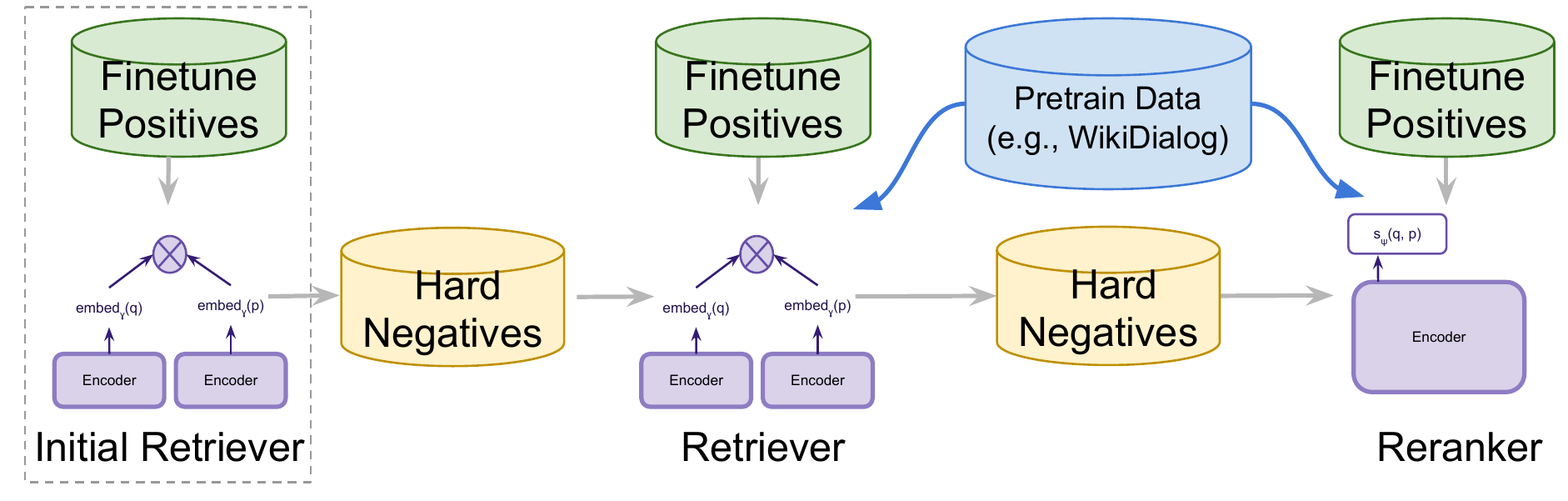}
\caption{\label{fig:retrieval-train}
Our pipeline for training conversational retrieval system. It follows the standard multi-stage training scheme used in previous work~\cite{lin2021TCT}.  We first train an initial retriever with in-batch negatives. We then train a second retriever on hard negatives from the initial retriever. Finally, we train a reranker on hard negatives from $\text{Retriever}$. 
Performance is further improved by pre-training Retriever and Reranker on document derived-dialogs.  }
\end{figure}

A training example from a typical retrieval dataset consists of a query paired with a positive passage, $(q, p^*)$. However, negative passages $\mathcal{N}(q)$ are usually not provided. Hence, we need to generate our own. We use two types of commonly used negatives: in-batch negatives , and  ``hard`` negatives mined with a multi-stage training strategy. 

\textbf{In-batch negatives}. When training with a batch, we treat the positive passage for example $i$ as a negative for all other examples $\neq i$ in the same batch. 

\textbf{``Hard'' negatives}.  Previous work has identified several weakness of in-batch negatives and identified the importance of selecting challenging negative examples for training retrievers~\cite{Qu2021RocketQA, Xiong2021ANCE, Santhanam2021ColBERTV2, lin2021TCT}.
Hence, following prior work~\cite{lin2021TCT}, we adopt a multi-stage training strategy where new negatives are mined at each stage (~\reffig{retrieval-train}):
\begin{enumerate}
    \item We train an initial retriever using only positives and in-batch negatives.
    \item We run top-$K$ retrieval using the initial retriever, and then randomly sample a subset of those to serve as negatives (when $K$ is large, a high percentage of these are true negatives).
    The hard negatives are then combined with positives to train a second retriever.
    \item We use the second retriever to again generate hard negatives. We train our reranker on these hard negatives combined with any positives retrieved by the second retriever.
\end{enumerate}
When pre-training on \WikiDialog{}, we use in-batch negatives. When fine-tuning for a downstream task, we use the multi-stage hard negative strategy.

Note that at inference time, we use the second retriever for top-K retrieval, and use the reranker to refine the top-K ranking (the initial retriever is not used at inference time). 
\section{Detailed Experimental Setup}
\label{sec:appendix:evaluation}

\subsection{Datasets}

\begin{table}[t]
\small
\setlength{\tabcolsep}{2pt}
\begin{tabular}{lcccc}
\toprule
                                  & \multicolumn{1}{l}{\textbf{\orquac{}}} & \multicolumn{1}{l}{\textbf{\qrecc{}}} & \multicolumn{1}{l}{\textbf{CAsT-19}} & \multicolumn{1}{l}{\textbf{CAsT-20}} \\ \midrule
\# train/dev Dialogs  & 4k/0.5k               & 10k/0.5k          & 0                           & 0  \\
\# train/dev Ques.     & 31k/3.4k          & 28k/1.5k        & 0                          & 0                          \\
\# test Dialogs   & 771               & 774          & 20                           & 25  \\
\# test Ques.     & 5571          & 8209        & 173                          & 208                          \\
\# Doc.                       & 11M                         & 54M                       & 38M & 38M \\                  
\bottomrule
\end{tabular}
\caption{\label{tab:retrieval-datasets} Statistics of  ConvQA Retrieval datasets.\cast{}-19 have 30 train and 50 test dialogs in total, but only 173 questions from 20 test dialogs have the relavance labels for evaluating retrieval model. Similarly, \cast{}-20 only has 208 questions from 25 dialogs with relevance labels. }
\end{table}

We use four open-domain conversational QA retrieval benchmarks: OR-QuAC~\cite{Qu2020ORQuAC},   TREC CAsT-19~\cite{ByrneKSNGDYDKC19},  TREC CAsT-20~\cite{Dalton2020TRECCAst20}, and QReCC~\cite{Anantha2021QReCC}. Table~\ref{tab:retrieval-datasets} lists their statistics.

\orquac{} and \qrecc{} were introduced in \refapp{appendix:train_dataset}.

TREC CAsT-19 and CAsT-20 are two datasets from the The TREC Conversational Assistance Track (CAsT)
shared task~\cite{Dalton2019TRECCAst19, Dalton2020TRECCAst20} with small numbers of dialogues for evaluating information-seeking conversational search systems. Questions in a dialogue are constructed manually
to mimic a “real” dialogue on a certain topic. The retrieval corpus includes web passages from MS MARCO~\cite{Nguyen2016MSMARCO} and
wikipedia passages from TREC Complex Answer Retrieval (CAR)~\cite{Dusart2019TRECCAR}. CAsT-19 provides human relevance labels for 173 questions in 20 test dialogues. CAsT-20 provides human relevance labels for 208 questions in 25 test dialogues.  

Note, \qrecc{} reuses question sequences from QuAC and TREC CAsT-19~\cite{ByrneKSNGDYDKC19}. However, although the questions are the same as existing datasets, \emph{the relevant passages and answers are different}, as it asked human raters to retrieve passages with a search engine and generate answers. In addition, TREC questions only appear in the \emph{test} set.  

In \orquac{}, at least 99.5\% of dialogs contain answers from the same answer passage, while in \cast{} and \qrecc{}, each question turn can be answered by a different passage. 

We follow the official \emph{automatic} setting~\cite{ByrneKSNGDYDKC19, Dalton2020TRECCAst20} for both datasets, which only uses the questions as retrieval inputs. We follow the authors of QReCC~\cite{Anantha2021QReCC} and \cite{Wu2021CONQRR} and use the both questions and gold answers from conversation history as retrieval inputs.

\subsection{Implementation}

 We implement dual-encoder retrievers and rerankers in JAX. 
 
\paragraph{Retrievers.} For pre-training on our inpainted datasets, we used a softmax temperature $\tau$ of 0.01, batch size 2048, and dropout rate 0.1. The models were trained with Adafactor optimizer with learning rate $1e^{-3}$ and 1k warm up steps. For checkpoint selection, we tested checkpoints at 50k and 100k steps and  reported the better one based on each finetune datasets' dev set performance. we fine-tuned the retrievers for 500 steps on \orquac{} and \qrecc{}. We did not fine-tune the retrievers on TREC CAsT-19 and CAsT-20 due to the small data size. We report TREC performance by retrieving with a QReCC retriever (trained without answers in the inputs). 

Questions and passages are always lowercased. Maximum query length was set to 128 for all pretrain and fine-tune datasets except for \qrecc{}, which uses 512 because it allows using previous answers in the queries. Maximum passage length was set to 256.

\textbf{Rerankers.} Rerankers were implemented using T5 encoder-decoder architecture. We pre-trained the rerankers on synthetic dialogs for 1.7M steps.   For fine-tuning on \orquac{} and \qrecc{}, we used 10 hard negatives for each question sampled from the top 100 passages returned by  the retriever. Fir TREC CaST-19 and Cast-20, we follow prior work~\cite{Yu2021ConvDR} and used 5-fold cross-validation to fine-tune the reranker\footnote{Theoretically, one could also use cross-validation to fine-tune retrievers on TREC CaST. However, it is very expansive due to inference on the retrieval corpus, so we did not fine-tune retriever in this work.}. We used 20 hard negatives per question randomly sampled from the top 100 retrieved results. We warm up the TREC rerankers on \qrecc{}. 

All reranker training used the Adafactor optimizer with constant learning rate $1e^{-3}$ and dropout rate 0.1. We used batch size 512 for pre-training, 128 for fine-tuning of \orquac{} and \qrecc{}, and 32 for fine-tuning of TREC CaST-19 and CaST-20. 




\subsection{Published baselines}
\label{sec:appendix:baselines}

We include five published \emph{retreival-only} baselines to compare with our DI retrievers. 

BM25-Query Rewriter~\cite{Yu2021ConvDR} and BM25-T5QR~\cite{Wu2021CONQRR} are two query rewriting approaches that trains a model to rewrite the dialog history into a contextualized, keyword-like query.  The former trains a GPT-2 query rewriter on ad hoc search sessions~\cite{Yu2020FewShotQueryRewrite}.  The latter  trains a T5 query rewriter on human-generated query rewrites from \qrecc{}.  Both systems issue the rewritten query to a classic BM25 lexical retrieval system.

ANCE-Query Rewriter~\cite{Yu2021ConvDR} uses the GPT-2 query rewriter from~\cite{Yu2021ConvDR}, but it runs the rewritten query with ANCE~\cite{Xiong2021ANCE}, is a strong non-conversational dense retriever trained on \msmarco{}. 

CONQRR~\cite{Wu2021CONQRR} is the previous state-of-the-art retrieval system on \qrecc{}. CONQRR uses reinforcement learning to optimize the query rewriter for the retriever.Its retriever is a t5-base dual-encoder trained on \msmarco{}.

ConvDR~\cite{Yu2021ConvDR} is the previous state-of-the-art conversational dense retrieval system on \orquac{} and TREC CaST tasks. Unlike the above approaches which all use query rewriting, ConvDR learns a conversational query encoder that directly maps the entire dialog history into a dense embedding. To address the bottleneck of limited conversational training data, ConvDR uses a teacher-student framework that trains
the student conversational query encoder to “mimic” the representation of the oracle query rewrite from a non-conversational teacher.

In addition to the retreval baselines, we include three published state-of-the-art conversational search systems that uses the \emph{retrieval+reranker} pipeline,  serving as baselines for our DI retriever + reranker system. 

CFDA\_CLIP\_RUN7~\cite{Yang2019CFDA} is the best performing system participated in TREC CAsT-19 automatic setting~\cite{Dalton2019TRECCAst19}. It uses heuristics to expand queries, BM25 with T5 doc2query for retrieval, and BERT to rerank.

h2oloo\_RUN4 is the best participating system in TREC CAsT-20 automatic setting~\cite{Dalton2020TRECCAst20}. It uses a T5 model for query rewriting, a dense-sparse hybrid retriever, and a T5 reranker trained on MS MARCO.  

ConvDR$\rightarrow$BERT~\cite{Yu2021ConvDR} uses a BERT reranker to rerank ConvDR~\cite{Yu2021ConvDR}'s retrieval results. The BERT reranker is pre-trained on MS MARCO and then fine-tuned for specific datasets on both the ranking loss and the query rewrite distillation loss.

\onecolumn

\section{Additional Experimental Results}

~\reftab{main-results-appendix} reports additional retrieval metrics in addition to the MRR reported in \reftab{main-results}. Specifically, we report recall and mean reciprocal rank at rank 5 (R@5 and MRR@5) following previous work~\citet{Qu2020ORQuAC, Yu2021ConvDR}.  On \qrecc{}, we report recall at rank 10 (R@10) and mean reciprocal rank without rank cut off (MRR) following~\citet{Anantha2021QReCC, Wu2021CONQRR}. On TREC CAsT-19 and CAsT-20, we use the official metrics MRR and NDCG@3 suggested by ~\citet{Dalton2019TRECCAst19, Dalton2020TRECCAst20}. Note that TREC CAsT-19 uses relevant grade $\geq1$ as positive for MRR but TREC CAsT-20 uses relevance grade $\geq2$ as positive for MRR~\cite{Dalton2020TRECCAst20}

\label{sec:appendix:results}
\begin{table}[H]
\centering
\renewcommand{\arraystretch}{0.85}
\small
\begin{tabular}{l rl rl rl rl}

\toprule
            & \multicolumn{2}{c}{\qrecc{}} & \multicolumn{2}{c}{\orquac{}} & \multicolumn{2}{c}{TREC CAsT-19} & \multicolumn{2}{c}{TREC CAsT-20} \\ 
            \cmidrule(lr){2-3}   \cmidrule(lr){4-5}      \cmidrule(lr){6-9}  
\textbf{System} 
    & \small MRR   & \small R@10 
    & \small MRR@5 & \small R@5 
    & \small MRR   & \small NDCG@3
    & \small MRR   & \small NDCG@3 
    \\ \midrule 
& \multicolumn{8}{c}{\textit{Retrieval}}                         \\
BM25-QR     & --   & --   & 20.2 & 30.2 & 58.1 & 27.7 & 25.0 & 15.9   \\
ANCE-QR     & --   & --   & 45.7 & 58.4 & 66.5 & 40.9 & 37.5 & 25.5   \\
ConvDR      & --   & --   & 61.6 & 75.0 & 74.0 & 46.6 & 50.1 & 34.0   \\
BM25-T5QR   & 32.8 & 52.5 & --   & --   & --   & --   & --   & --    \\
CONQRR      & 41.8 & 65.1 & --   & --   & --   & --   & --   & --    \\  \cmidrule(l){1-1}
T5-Large DE & 55.7 & 78.7 & 56.9 & 69.7 & 61.0 & 32.5 & 34.3 & 19.9 \\ 
\quad $\triangleright$ \small{\textsc{Wiki}}
            & 60.4 & \textbf{83.3} & 66.5 & 77.9 & 68.1 & 40.3 & 43.7 & 27.9 \\   
\quad $\triangleright$ \textbf{\small{\textsc{Wiki+Web}}} 
            & \textbf{60.7} & 83.1 & \textbf{68.7} & \textbf{80.3} & \textbf{74.1} & \textbf{47.0} & \textbf{51.3} & 33.2  \\ \midrule \midrule
& \multicolumn{8}{c}{\textit{Retrieval + Reranking}}                  \\ 
CFDA\_CLIP\_RUN7 
            & --   & --  & --   & --   & 71.4 & 43.6 & --   & -- \\
h2oloo\_RUN4
            & --   & --  & --   & --   & --   & --   & 59.3 & 45.8  \\
ConvDR$\rightarrow$BERT 
            & --   & --     & 77.3 & 85.0 & 79.9 & 54.1 & 54.5 & 39.2   \\ \cmidrule(l){1-1}
T5-Large DE  $\rightarrow$ reranker
            & 68.9 & 86.0   & 72.6 & 78.6 & 75.3 & 48.3 & 55.1 & 36.7 \\
\quad $\triangleright$ \small{\textsc{Wiki}} 
            & 70.7 & 88.8   & 79.7 & 86.8 &  79.3 & 50.7 & \textbf{60.3} & 43.0 \\
\quad $\triangleright$ \textbf{\small{\textsc{Wiki+Web}}} 
            & \textbf{71.8} & \textbf{88.9}   & \textbf{81.2} & \textbf{87.9} & \textbf{82.0} & 53.6 & 59.7 & 42.5 \\
\bottomrule
\end{tabular}
\caption{\label{tab:main-results-appendix} Full evaluation on \qrecc{}, \orquac{} and TREC \cast{}.  This table reports additional metrics for these datasets following prior work~
\citep{Yu2021ConvDR, Wu2021CONQRR} in addition to MRR reported in \reftab{main-results}. 
} \
\end{table}
\section{Inpainted Dialog Examples}
\label{sec:appendix:examples}

Tables \ref{tab:example-1}-\ref{tab:example-4} show additional examples of inpainted dialogs.

\begin{table}
\centering
\begin{tabular}{l|p{14cm}} 
\multicolumn{2}{c}{Article: European School, Munich} \\ \toprule
\InpaintST{} & 
1: What is the European School, Munich?

0: The European School, Munich (ESM) is one of thirteen European Schools and one of three in Germany.

1: What did the European School, Munich do?

0: First established in 1977, it moved to its current location in Neuperlach, a district in the south-east of Munich, in the state of Bavaria in 1981.

1: Are there any other interesting aspects about this article?

0: The ESM was principally established to serve the schooling needs of children of the staff of the European Patent Office (EPO) – the executive body of the European Patent Organisation.

1: Did it serve any other students?

0: However, enrolment is open to other prospective students.

1: What other services did it offer?

0: The school offers the European Baccalaureate as its secondary leaving qualification." \\ \midrule
\InpaintOQ{} &
1: What is the European School in Munich?

0: The European School, Munich (ESM) is one of thirteen European Schools and one of three in Germany.

1: Where is it located?

0: First established in 1977, it moved to its current location in Neuperlach, a district in the south-east of Munich, in the state of Bavaria in 1981.

1: Who attended/teaches there?

0: The ESM was principally established to serve the schooling needs of children of the staff of the European Patent Office (EPO) – the executive body of the European Patent Organisation.

1: Was it later expanded to serve the needs of the general public?

0: However, enrolment is open to other prospective students.

1: what types of classes are taught there?

0: The school offers the European Baccalaureate as its secondary leaving qualification." \\ \midrule
\InpaintSTOQ{} &
1: What is the purpose of the European School, Munich?

0: The European School, Munich (ESM) is one of thirteen European Schools and one of three in Germany.

1: Where is it located?

0: First established in 1977, it moved to its current location in Neuperlach, a district in the south-east of Munich, in the state of Bavaria in 1981.

1: Is it open to anyone?

0: The ESM was principally established to serve the schooling needs of children of the staff of the European Patent Office (EPO) – the executive body of the European Patent Organisation.

1: Are there any other interesting aspects about this article?

0: However, enrolment is open to other prospective students.

1: What are some of the courses taught?

0: The school offers the European Baccalaureate as its secondary leaving qualification."
\\ \bottomrule

\end{tabular}
\caption{\label{tab:example-1} Synthetic dialogs about \emph{European School, Munich} generated by \InpaintST{},\InpaintOQ{}, and \InpaintSTOQ{}.}
\end{table}

\begin{table*}
\centering
\begin{tabular}{l|p{14cm}} 
\multicolumn{2}{c}{Article: Ageing, Disability and Home Care NSW} \\ \toprule
\InpaintST{} & 
1: What is Ageing, Disability and Home Care NSW?

0: Ageing, Disability and Home Care NSW (ADHC) is a division of the Department of Family and Community Services in the Government of New South Wales that is responsible for the provision of services to older people, people with a disability, and their families and carers in the state of New South Wales, Australia.

1: What services does it provide?

0: The ADHC division supports approximately 300,000 vulnerable people from its annual budget of 2.5 billion.

1: Does it have a large staff?

0: It employs more than 13,000 staff.

1: What is it known for?

0: ADHC delivers services and manages 900 non-government and local government service providers that are funded to deliver services such as early intervention, therapy, respite, supported accommodation, case management and post-school programs.

1: Who runs it?

0: The division is led by a Deputy Secretary, presently Jim Longley, who reports to the Minister for Ageing and Minister for Disability Services, presently the Hon. John Ajaka MLC.
\\ \midrule
\InpaintOQ{} &
1: What is Ageing, Disability and Home Care NSW?

0: Ageing, Disability and Home Care NSW (ADHC) is a division of the Department of Family and Community Services in the Government of New South Wales that is responsible for the provision of services to older people, people with a disability, and their families and carers in the state of New South Wales, Australia.

1: How many people do they serve?

0: The ADHC division supports approximately 300,000 vulnerable people from its annual budget of 2.5 billion.

1: How many employees do they have?

0: It employs more than 13,000 staff.

1: What kind of services do they provide?

0: ADHC delivers services and manages 900 non-government and local government service providers that are funded to deliver services such as early intervention, therapy, respite, supported accommodation, case management and post-school programs.

1: Who are the executives?

0: The division is led by a Deputy Secretary, presently Jim Longley, who reports to the Minister for Ageing and Minister for Disability Services, presently the Hon. John Ajaka MLC.
\\ \midrule
\InpaintSTOQ{} &
1: Who does Ageing, Disability and Home Care NSW (ADHC) work for?

0: Ageing, Disability and Home Care NSW (ADHC) is a division of the Department of Family and Community Services in the Government of New South Wales that is responsible for the provision of services to older people, people with a disability, and their families and carers in the state of New South Wales, Australia.

1: How much does it cost?

0: The ADHC division supports approximately 300,000 vulnerable people from its annual budget of 2.5 billion.

1: How many people work for the division?

0: It employs more than 13,000 staff.

1: What services does the division provide?

0: ADHC delivers services and manages 900 non-government and local government service providers that are funded to deliver services such as early intervention, therapy, respite, supported accommodation, case management and post-school programs.

1: Who is in charge of the division?

0: The division is led by a Deputy Secretary, presently Jim Longley, who reports to the Minister for Ageing and Minister for Disability Services, presently the Hon. John Ajaka MLC.
\\ \bottomrule
\end{tabular}
\caption{\label{tab:example-2} Synthetic dialogs about \emph{Ageing, Disability and Home Care NSW} generated by \InpaintST{},\InpaintOQ{}, and \InpaintSTOQ{}.}
\end{table*}

\begin{table*}
\centering
\begin{tabular}{l|p{14cm}}
\multicolumn{2}{c}{Article: FAQ} \\ \toprule
\InpaintST{} & 
1: What is FAQ?

0: An FAQ is a list of frequently asked questions (FAQs) and answers on a particular topic (also known as Questions and Answers [Q\&A] or Frequently Answered Questions).

1: Are there any other interesting aspects about this article?

0: The format is often used in articles, websites, email lists, and online forums where common questions tend to recur, for example through posts or queries by new users related to common knowledge gaps.

1: Is there anything else important?

0: The purpose of an FAQ is generally to provide information on frequent questions or concerns; however, the format is a useful means of organizing information, and text consisting of questions and their answers may thus be called an FAQ regardless of whether the questions are actually "frequently" asked.

1: Any other information?

0: Since the acronym "FAQ" originated in textual media, its pronunciation varies.

1: How does it vary?

0: FAQ is most commonly pronounced as an initialism, "F-A-Q", but may also be pronounced as an acronym, "FAQ".
\\ \midrule
\InpaintOQ{} &
1: What is meant by a "Frequently asked question"?

0: An FAQ is a list of frequently asked questions (FAQs) and answers on a particular topic (also known as Questions and Answers [Q\&A] or Frequently Answered Questions).

1: Who asks the questions?

0: The format is often used in articles, websites, email lists, and online forums where common questions tend to recur, for example through posts or queries by new users related to common knowledge gaps.

1: Is the information intended for a specific audience?

0: The purpose of an FAQ is generally to provide information on frequent questions or concerns; however, the format is a useful means of organizing information, and text consisting of questions and their answers may thus be called an FAQ regardless of whether the questions are actually "frequently" asked.

1: How is the term "FAQ" pronounced?

0: Since the acronym "FAQ" originated in textual media, its pronunciation varies.

1: What does it vary in pronunciation?

0: FAQ is most commonly pronounced as an initialism, "F-A-Q", but may also be pronounced as an acronym, "FAQ".
\\ \midrule
\InpaintSTOQ{} &
1: what is the purpose of a faq

0: An FAQ is a list of frequently asked questions (FAQs) and answers on a particular topic (also known as Questions and Answers [Q\&A] or Frequently Answered Questions).

1: Are these often used on a website?

0: The format is often used in articles, websites, email lists, and online forums where common questions tend to recur, for example through posts or queries by new users related to common knowledge gaps.

1: What questions are asked most frequently?

0: The purpose of an FAQ is generally to provide information on frequent questions or concerns; however, the format is a useful means of organizing information, and text consisting of questions and their answers may thus be called an FAQ regardless of whether the questions are actually "frequently" asked.

1: Is FAQ pronounced as an acronym?

0: Since the acronym "FAQ" originated in textual media, its pronunciation varies.

1: How is it pronounced?

0: FAQ is most commonly pronounced as an initialism, "F-A-Q", but may also be pronounced as an acronym, "FAQ".
\\ \bottomrule
\end{tabular}
\caption{\label{tab:example-3} Synthetic dialogs about \emph{FAQ} generated by \InpaintST{},\InpaintOQ{}, and \InpaintSTOQ{}.}
\end{table*}

\begin{table*}
\centering
\begin{tabular}{l|p{14cm}}
\multicolumn{2}{c}{Article: George B. McClellan} \\ \toprule
\InpaintST{} & 
1: Did George B. McClellan subordinates blame him for the failures on the campaign of 1862?

0: Although McClellan's subordinates can claim their share of responsibility for delays (such as Ambrose Burnside's misadventures at Burnside Bridge) and blunders (Edwin V. Sumner's attack without reconnaissance), these were localized problems from which the full army could have recovered.

1: What other blunder happened during the campaign?

0: Historians have faulted McClellan for accepting the cautious advice about saving his reserves, such as when a significant breakthrough in the center of the Confederate line could have been exploited, but Fitz John Porter is said to have told McClellan, "Remember, General , I command the last reserve of the last Army of the Republic." However, the veracity of this supposed statement is in doubt.

1: Why is it in doubt?

0: Porter in the post-war period never claimed to have made the statement and it also fails to note the several Union corps at that time defending Washington.

1: Did the campaign end in a draw?

0: Despite being a tactical draw, Antietam is considered a turning point of the war and a victory for the Union because it ended Lee's strategic campaign (his first invasion of the North) and it allowed President Lincoln to issue the Emancipation Proclamation on September 22, taking effect on January 1, 1863.
\\ \midrule
\InpaintOQ{} &
1: What was George B. McClellan's biggest failure?

0: Although McClellan's subordinates can claim their share of responsibility for delays (such as Ambrose Burnside's misadventures at Burnside Bridge) and blunders (Edwin V. Sumner's attack without reconnaissance), these were localized problems from which the full army could have recovered.

1: What was his biggest success?

0: Historians have faulted McClellan for accepting the cautious advice about saving his reserves, such as when a significant breakthrough in the center of the Confederate line could have been exploited, but Fitz John Porter is said to have told McClellan, "Remember, General , I command the last reserve of the last Army of the Republic." However, the veracity of this supposed statement is in doubt.

1: Why is the veracity of this statement in doubt?

0: Porter in the post-war period never claimed to have made the statement and it also fails to note the several Union corps at that time defending Washington.

1: What was the biggest battle he was involved in?

0: Despite being a tactical draw, Antietam is considered a turning point of the war and a victory for the Union because it ended Lee's strategic campaign (his first invasion of the North) and it allowed President Lincoln to issue the Emancipation Proclamation on September 22, taking effect on January 1, 1863.
\\ \midrule
\InpaintSTOQ{} &
1: What did George B. McClellan do during the Seven Days' Campaigns?

0: Although McClellan's subordinates can claim their share of responsibility for delays (such as Ambrose Burnside's misadventures at Burnside Bridge) and blunders (Edwin V. Sumner's attack without reconnaissance), these were localized problems from which the full army could have recovered.

1: Are there any other interesting aspects about this article?

0: Historians have faulted McClellan for accepting the cautious advice about saving his reserves, such as when a significant breakthrough in the center of the Confederate line could have been exploited, but Fitz John Porter is said to have told McClellan, "Remember, General , I command the last reserve of the last Army of the Republic." However, the veracity of this supposed statement is in doubt.

1: Why is it in doubt?

0: Porter in the post-war period never claimed to have made the statement and it also fails to note the several Union corps at that time defending Washington.

1: Was it a successful campaign?

0: Despite being a tactical draw, Antietam is considered a turning point of the war and a victory for the Union because it ended Lee's strategic campaign (his first invasion of the North) and it allowed President Lincoln to issue the Emancipation Proclamation on September 22, taking effect on January 1, 1863.
\\ \bottomrule
\end{tabular}
\caption{\label{tab:example-4} Synthetic dialogs about \emph{George B. McClellan} generated by \InpaintST{},\InpaintOQ{}, and \InpaintSTOQ{}.}
\end{table*}
\section{Studying Sensitive Language in the Dataset}
\label{sec:appendix:safety}

Prior to evaluating the dataset for fairness implications, we found it necessary to define “fairness” in the context of the model’s generated queries. We began our inquiry by considering conversational fairness – that is, what constitutes  insensitive, offensive, or prejudiced speech in a dialogue? We then qualitatively assessed a small sample of the dataset and described the following potential fairness related concerns that may arise in the dataset:
\begin{itemize}
  \item Model generates a query that contains sensitive characteristics that were not explicitly mentioned in the passage.
  \item Model overemphasizes specific sensitive characteristics or identities in a negative or offensive way. For example: the model continues to generate questions related to a writer’s biracial identity when it is not the intended focus of the passage.
  \item Model ascribes negative values to sensitive characteristics without that being explicitly mentioned in the passage. For example, the model generates a question about “how barbaric” a society is for a passage about a non-Western society.
  \item Model produces “unhelpful” or repetitive questions when the source passage contains sensitive characteristics as compared to passages with non-sensitive characteristics.
\end{itemize}

The primary approach to evaluating fairness in the dataset involved scanning for sensitive terms and interactions between those terms. To this end, we curated a list of sensitive terms from the literature that includes nouns and adjectives that are considered harmful or insensitive when applied to particular groups (citation). When a combination of these terms appeared in a question in the dataset (\reftab{safety-category}), we considered it to be potentially sensitive. 
From this initial analysis, we observed that 0.2-0.5\% of the dataset contains such potentially sensitive interactions of terms in generated queries.
See Tables~\ref{tab:sensitivity-example-1}--\ref{tab:sensitivity-example-5} for selected examples that matched this filter.

\begin{table}[H]
\centering
\small
\begin{tabular}{l | c c c c c c} \toprule
                        & Race/Ethnicity    & Nationality   & Gender        & Region    & Political     & Religion \\ \midrule
Adjective               & \checkmark        & \checkmark    & \checkmark    &           &               &          \\
Violence Related        & \checkmark        & \checkmark    &               & \checkmark&               &\checkmark \\
Competence Adjective    & \checkmark        & \checkmark    & \checkmark    &           &               & \checkmark    \\
Physical Appearance Adjectives  & \checkmark    &           & \checkmark    &           &               &          \\
Outsider Adjectives     & \checkmark        & \checkmark    & \checkmark    &           &               & \checkmark    \\
Harmful Racial Stereotypes  & \checkmark    &               & \checkmark    &           &               &          \\
Race/Ethnicity Adjectives   & \checkmark    & \checkmark    &               &           &               &          \\
Gender Adjectives       & \checkmark        &               & \checkmark    &           &               &          \\
Political               &                   &               &               & \checkmark&               &          \\
Sexual Orientation      & \checkmark        &               & \checkmark    &           &               &          \\
Socio-economic/class    & \checkmark        & \checkmark    &               &           &\checkmark     &          \\
Occupation              & \checkmark        & \checkmark    & \checkmark    &           &               &          \\ \bottomrule
\end{tabular}
\caption{\label{tab:safety-category} The categories of sensitive terms and their interactions.}
\end{table}

\begin{table}[H]
    \centering
    \begin{tabular}{p{16cm}} \toprule
         \multicolumn{1}{c}{Article: Model minority}\\ \midrule
1: The term model minority is often used in reference to \textbf{African} Americans, \textbf{Hispanic} Americans, and Asian Americans. But what does the term actually mean?

0: The model minority label also includes South Asian communities, in particular, Indian Americans, because of their high socioeconomic success.

1: And what can Indian Americans do to help other Asian Americans?

0: But, Indian-Asians face a type of racism and discrimination, despite their Model Minority status, that other Asians do not face.

1: Can you tell me more about this racism?

0: One of the forms of this is discrimination based on religion; often times, Indians, specifically Sikhs, in America face racism that that is a result of mislabeling like being called Middle-Eastern or Muslim by the way they look, whether they actually identify with those groups, with Muslim Americans being markedly poorer than Hindu Americans.

1: And how does this impact other Asian Americans?

0: The mass grouping of all different Asian ethnicities into one ``positive'' stereotype delegitimizes the problems that certain Asian immigrants face, because the majority might not face those same negative stereotypes.

1: What can the Sikhs do to help other Asian immigrants?

0: Another result of the Model Minority status of Indian-Asians is their complacency, expressed more often than not.

1: How does this affect other Asian immigrants?

0: There is conflict within the Indian-Asian immigrant population in the US, as some exalt their ``positive'' minority status and wish to uphold it, resulting in complacency when attacked racially and even sometimes going as far as to support political candidates that are anti-immigration. 
\\ \bottomrule
    \end{tabular}
    \caption{\label{tab:sensitivity-example-1} This dialog was flagged as potentially sensitive because it contained sensitive terms ``african'' and ``hispanic''.}
\end{table}

\begin{table}[H]
    \centering
    \begin{tabular}{p{16cm}} \toprule
         \multicolumn{1}{c}{Article: The Queers}\\ \midrule
1: What was the band The Queers' response to the November 2015 Woolsey Fire disaster?

0: In an interview with Noisey, after their February 2016 show was cancelled due to a call for a petition to boycott the show from a local collective, Babely Shades, Joe Queer voiced his opinion when asked if he ``supports racist, misogynistic or homophobic values?'' he replied ``Of course not.''

1: Did Joe Queer think that the Woolsey Fire was man-made?

0: ``What normal minded person does?''

1: Did Joe Queer have any black friends?

0: ``My closest pals here in Atlanta are Chris and Chester, two gay black guys.''

1: Was Joe Queer in school when the Woolsey Fire happened?

0: ``I was actually working a part time job.''

1: Was Joe Queer working with any black people at the time?

0: ``I was one of two white people on the whole job.''

1: Did Joe Queer think that \textbf{African Americans} were \textbf{lazy} and lacked work ethic?

0: ``If I was so racist, I wouldn't have worked there or be living in Atlanta!''
\\ \bottomrule
    \end{tabular}
    \caption{\label{tab:sensitivity-example-2} This dialog was flagged as potentially sensitive because it contains sensitive terms ``african'' and ``lazy''.}
\end{table}

\begin{table}[H]
    \centering
    \begin{tabular}{p{16cm}} \toprule
         \multicolumn{1}{c}{Article: Female intrasexual competition}\\ \midrule
1: How does a woman's ovulation cycle affect the way she competes with other women?

0: This indirect form of competition appears exclusive toward females as findings also showed that women, irrespective of ovulation status (high or low), showed no difference in the rating of male faces.

1: What other behaviors or factors affect the way women compete with each other?

0: Supporting research has also found that younger women who are considered as having high fertility, gossip about other women more than older women, who are no longer at their most fertile stage.

1: Do women tend to be more \textbf{aggressive} toward each other than \textbf{men}?

0: Indeed, indirect aggression appears more prevalent amongst (or exclusive to) females than males who are said to engage in more direct forms of competition.

1: Are there any other interesting aspects about this article?

0: Research studying the relationship between indicators of attractiveness, such as physical attractiveness and indirect victimisation, showed that the likelihood of experiencing indirect victimization increased by 35\% for females who perceived themselves as physically attractive.

1: Is this something that also happens with men?

0: In contrast, being a male who is physically attractive decreased the chances of experiencing such indirect victimization.

1: Why is this difference in behavior between men and women important?

0: This also highlights how the physical attractiveness a female is a trigger for indirect aggression and forms a core part of intersexual selection between the sexes. \\ 
\bottomrule
    \end{tabular}
    \caption{\label{tab:sensitivity-example-3}
    The dialog is potentially sensitive due to containing sensitive terms ``aggressive'' and ``men''.}
\end{table}

\begin{table}[H]
    \centering
    \begin{tabular}{p{16cm}} \toprule
         \multicolumn{1}{c}{Article: Virginia M. Alexander}\\ \midrule
1: who was the first \textbf{african american} woman to become a \textbf{doctor} in the united states? \\
0: Virginia M. Alexander  Virginia M. Alexander (February 4, 1899 – July 24, 1949) was an American physician, public health researcher, and the founder of the Aspiranto Health Home in Philadelphia, Pennsylvania. \\
1: what was her childhood like? \\
0: Virginia M. Alexander was born in Philadelphia, Pennsylvania on February 4, 1899 to Hilliard Alexander and Virginia Pace. \\
1: did she have siblings? \\
0: She had four siblings, including the prominent attorney Raymond Pace Alexander. \\
1: what were her parents like? \\
0: Alexander's mother died when she was 4 years old, and at age 13 her father's riding academy closed. \\
1: what did she do after her parents' academy closed? \\
0: Alexander withdrew from school to help relieve the resulting economic strain on her family, but her father insisted that she finish her education. \\
1: where did she go to school? \\
0: Alexander attended high school at the William Penn High School for Girls, where she graduated with honors before receiving a scholarship that allowed her to attend the University of Pennsylvania to complete her undergraduate education. \\
\bottomrule
    \end{tabular}
    \caption{\label{tab:sensitivity-example-4}
    This dialog is flagged as potentially sensitive because it contains the sensitive terms ``African-American'' and ``doctor''. The first answer is not factually accurate---Virginia M. Alexander is not the first African-American physician.}
\end{table}

\begin{table}[H]
    \centering
    \begin{tabular}{p{16cm}} \toprule
         \multicolumn{1}{c}{Article: Symbolism of terrorism}\\ \midrule
1: ``\textbf{Allah} is most gracious to the oppressed (Quran passage)'' is mentioned in the analysis of a terrorist \textbf{attack}. Is this quote from the Quran a part of the analysis? \\
0: The analysis does not discuss the quote on the Quran. \\
1: What is the image in the analysis? \\
0: The central figure of the image is a yellow tiger, which is surrounded by a ring of bullets and crossed by two rifles. \\
1: What is the significance of the image? \\
0: The script both around the tiger, in Tamil, and below the tiger, in English, reads ``Liberation Tigers of Tamil Eelam.'' Eelam is Tamil for the Island of Sri Lanka, which the groups seeks to liberate from the control of the Sinhalese majority. \\
1: What is the significance of the tiger? \\
0: The tiger is a culturally important animal that symbolizes heroism, militancy, and patriotism. \\
1: What is the significance of the tiger in this image? \\
0: Its centrality in this image indicates the desire of the Liberation Tigers to be self-determined and independent. \\
1: What is the significance of the bullets and rifles? \\
0: The bullets and rifles indicate their employment of violence as the means to achieve these goals. \\
\bottomrule
    \end{tabular}
    \caption{\label{tab:sensitivity-example-5}
    This dialog is flagged as potentially sensitive because it contains the sensitive terms ``allah'' and ``attack''.}
\end{table}

\end{document}